\documentclass[journal]{IEEEtran}

\usepackage{color}
\usepackage[table,usenames,dvipsnames]{xcolor}
\usepackage{graphicx}

\usepackage{amsmath}
\usepackage{mathtools}

\usepackage{cite}

\definecolor{linkcolor}{RGB}{0,0,128}

\usepackage[
            pdfpagelabels,
            plainpages=false,
            hypertexnames=true,
            colorlinks=true,
            linkcolor=linkcolor,
            anchorcolor=linkcolor,
            citecolor=linkcolor,
            filecolor=linkcolor,
            menucolor=linkcolor,
            runcolor=linkcolor,
            urlcolor=linkcolor,
            pdfborder={0 0 0} ,
            hidelinks]{hyperref}

\usepackage{dblfloatfix}

\usepackage{xargs} 

\usepackage{algpseudocode}
\usepackage{algorithm}

\makeatletter
\let\myorg@bibitem\bibitem
\def\bibitem#1#2\par{%
  \@ifundefined{bibitem@#1}{%
    \myorg@bibitem{#1}#2\par
  }{%
    \begingroup
      \color{\csname bibitem@#1\endcsname}%
      \myorg@bibitem{#1}#2\par
    \endgroup
  }%
}
\newcommand*{\bibitem@Slice}{blue}
\newcommand*{\bibitem@ARS}{blue}
\newcommand*{\bibitem@ARMS}{blue}
\newcommand*{\bibitem@panmarginal}{blue}
\makeatother

\newcommand{\makeblue}[1]{\textcolor{black}{#1}} 

\begin{document}
%
\title{Fuzzy Bayesian Learning}

%
%
%


\author{Indranil~Pan
        and~Dirk Bester

\thanks{I. Pan was with Imperial College London, UK, and now works at Sciemus Ltd, London UK. e-mail: i.pan11@imperial.ac.uk, ipan@sciemus.com}
\thanks{D. Bester was with University of Oxford, UK, and now works at Sciemus Ltd, London UK. e-mail: dbester@sciemus.com}}
\maketitle

\begin{abstract}
In this paper we propose a novel approach for learning from data using rule based fuzzy inference systems where the model parameters are estimated using Bayesian inference and Markov Chain Monte Carlo (MCMC) techniques. We show the applicability of the method for regression and classification tasks using synthetic data-sets and also a real world example in the financial services industry.
Then we demonstrate how the method can be extended for knowledge extraction to select the individual rules in a Bayesian way which best explains the given data. 
Finally we discuss the advantages and pitfalls of using this method over state-of-the-art techniques and highlight the specific class of problems where this would be useful.

\end{abstract}

\begin{IEEEkeywords}
fuzzy logic; mamdani method; machine learning; MCMC.  
\end{IEEEkeywords}

%
\IEEEpeerreviewmaketitle

\section{Introduction}
%
%
%
%

\IEEEPARstart{P}{robability} theory and fuzzy logic have been shown to be complementary \cite{zadeh1995discussion} and various works have looked at the symbiotic integration of these two paradigms \cite{YAGER1979113,ZADEH1984363} including the recently introduced concept of Z-numbers \cite{Zadeh20112923}. Historically fuzzy logic has been applied to problems involving imprecision in linguistic variables, while probability theory has been used for quantifying uncertainty in a wide range of disciplines. Various generalisations and extensions of fuzzy sets have been proposed to incorporate uncertainty and vagueness which arise from multiple sources. For example, the type-2 fuzzy \cite{dubois1980fuzzy}, \cite{mizumoto1976some} sets and type-n fuzzy sets \cite{dubois1980fuzzy} can include uncertainty while defining the membership functions themselves.
Intuitionistic fuzzy sets \cite{atanassov1986intuitionistic} additionally introduce the degree of non-membership of an element to take into account that there might be some hesitation degree and the degree of membership and non-membership of an element might not always add to one. Non-stationary fuzzy sets \cite{garibaldi2008nonstationary}  can model variation of opinion over time by defining a collection of type 1 fuzzy sets and an explicit relationship between them. Fuzzy multi-sets \cite{yager1986theory} generalise crisp sets where multiple occurrences of an element are permitted. Hesitant fuzzy sets \cite{torra2010hesitant} have been proposed from the motivation that the problem of assigning a degree of membership to an element is not because of a margin of error (like Atanassov's intuitionistic fuzzy sets) or a possibility distribution on possibility values (e.g. type 2 fuzzy sets), but because there is a set of possible values \cite{torra2010hesitant}. Formally these can be viewed as fuzzy multi-sets but with a different interpretation. Apart from these quantitative tools, other approaches have been used to address the vagueness of the qualitative meanings used by experts like fuzzy linguistic approaches \cite{zadeh1975concept} and various linguistic models have been proposed to address this \cite{dong2009computing, herrera20002, rodriguez2012hesitant}.

Bayesian inference, on the other hand, provides a method of incorporating the subjective beliefs of the experts about the value of a model parameter, using prior probability distributions, through expert elicitation methods \cite{gelman2014bayesian}. However, expert knowledge does not always come in the form of precise probability distributions and often have vague and ambiguous statements embodying some form of knowledge \cite{li2013dealing}. In such cases, both uncertainty and imprecision exist simultaneously \cite{rajabi2016efficient}. The data is said to be uncertain if the degree of confidence in what is represented by the data is less than one \cite{rajabi2016efficient,khaleghi2013multisensor}. On the other hand data is imprecise if the implied attribute is not singular but belongs to an ill defined set \cite{rajabi2016efficient,khaleghi2013multisensor}. These two things cannot be discerned if they are represented by a single probability distribution and the contribution of each of these in the final posterior distribution cannot be tracked \cite{ross2009aleatoric}.

However, fewer works have looked at integration of Bayesian methods with  fuzzy logic \cite{Viertl1987, Viertl2008}. For example, \cite{fruhwirth1993fuzzy} attempts to generalise Bayesian methods for samples of fuzzy data and for prior distributions with imprecise parameters. Recent works have also looked at extension of these works for possibilistic Bayesian inference \cite{arefi2010possibilistic}. However, in all the above-mentioned literatures, most of the work is on representing the data in terms of fuzzy numbers and optimising model parameters based on that. None of the literatures specifically leverage on one of the fundamental strengths of the fuzzy inference system, i.e. expressing the expert opinion in terms of rule bases. This is specifically what this paper addresses and the idea stems from the fact that often the experts not only have an opinion on the uncertainty estimates of the model covariates, but also have an intuition of the inter-dependence between the various model covariates on the final output. None of the above-mentioned methods take this explicitly into account and hence do not fully exploit the embodied knowledge provided by the experts. 

Another line of research has been the use of the belief rule base (BRB) inference methodology as proposed in \cite{yang2006belief} which uses an expert rule base and assigns a belief degree to each rule, individual rule weights and attribute weights. This treatment is not fully Bayesian and including these additional parameters for characterising these rule bases requires multi-objective optimisation methods to give acceptable levels of performance \cite{yang2007optimization}. Having to estimate more number of parameters from a small data-set is inherently difficult and would have large confidence intervals (using frequentist methods) or large highest density intervals (using Bayesian methods). This issue is addressed in our methodology, since it does not need to have too many additional parameters over and above that required by the fuzzy rule base itself. This also helps in reducing the explosion of parameters if higher number of rule bases or fuzzy sets are used. Also the outputs in these works is a belief distribution over the fuzzy variables like low, medium and high, instead of a distribution on the crisp output of real numbers, as proposed later in this work. This is important especially if such models are developed as a part of a larger decision making framework which relies on probability distributions as outputs from the sub-models. A Bayesian extension of this framework is proposed in \cite{zhou2011bayesian} to update the belief degrees based on a recursive online estimation. This also addresses a different problem and has the same issues as the BRB methodology mentioned earlier.  

Adaptive fuzzy systems have been used to approximate non-conjugate priors, likelihoods and approximate hyper-priors in Bayesian inference, which helps in allowing users to extend the inference for a wider range of prior and likelihood probability density functions \cite{osoba2011bayesian}. In our proposed method, we use Markov Chain Monte Carlo (MCMC) \cite{MCMCpracCh1} to numerically approximate the posterior distribution and hence do not have the issues relating to conjugacy. Consequently, our method can work on any likelihood or prior distribution.

The motivation of the present work stems from the need to incorporate linguistic expert opinion within the standard Bayesian framework, so that the end-to-end overall system works on a set of crisp input-output data, which might have probabilistic uncertainties associated with them. It is easier for experts with domain knowledge to come up with a linguistic rule base, but to get good prediction accuracy on any data-set, the shape of the membership functions need to be appropriately chosen. The novelty of our work is that we convert this problem into a Bayesian parameter estimation problem where the shape of the membership functions are estimated through an MCMC algorithm for a given data-set. We also extend this to Bayesian inference on the individual rules themselves to find out which of the rules are more probable, given the data.

This approach leverages on the strengths of both the disciplines -- statistics and computational intelligence to come up with a practical solution methodology. In traditional statistics, the emphasis is on parameter estimation for a given model structure (which is chosen by the user) from a given set of data-points. In soft computing, the emphasis is on mathematically laying down the imprecision in the spoken language and expert opinions in the form of rule bases and does not need data to generate membership functions at the outset (data might be used later for calibration of the membership functions to improve prediction accuracies). Both of these advantages are synergistically used in the present work and the subsequent section develops this proposition in more detail.  

\vspace{1ex}

\section{Proposed methodology and demonstrative examples}

All empirical modelling approaches which learn from data, like linear and non-linear regression, generalised linear models, neural networks, or kernel based methods like Gaussian processes, support vector machines etc. can be interpreted using a generic Bayesian framework as an inference of the underlying non-linear model function $g(\boldsymbol{x};\theta)$ (with parameters $\theta$), given the data. For the set of input vectors 
$\boldsymbol{X}_{N}=\boldsymbol{x}_{i}  \forall i \in \left\{ 1,2,...,N \right\}$
and the set of outputs 
$\boldsymbol{Y}_{N}=\boldsymbol{y}_{i}  \forall i \in \left\{ 1,2,...,N \right\}$
the inference of the function $g(\boldsymbol{x};\theta)$ can be mathematically represented by the posterior probability distribution \cite{mackay2003information}
\begin{equation}
\label{eq:myeqdef1}
p\left( g(\boldsymbol{x};\theta)\mid \boldsymbol{Y}_{N}, \boldsymbol{X}_{N} \right)
=\frac{
        p\left( \boldsymbol{Y}_{N}\mid g(\boldsymbol{x};\theta), \boldsymbol{X}_{N} \right)
        p\left( g(\boldsymbol{x};\theta) \right)
      }{
        p\left( \boldsymbol{Y}_{N} \mid \boldsymbol{X}_{N} \right)
       }
\end{equation}
where,
$p\left( \boldsymbol{Y}_{N}\mid g(\boldsymbol{x};\theta), \boldsymbol{X}_{N} \right)$
is the likelihood function which gives the probability of the output values given the function 
$g(\boldsymbol{x};\theta)$
and the term 
$p\left( g(\boldsymbol{x};\theta) \right)$
is the prior distribution of the functions and its corresponding parameters assumed by the model.
As with most other modelling approaches, the above equation assumes that the covariates are independent of the model parameters and the functional form of $g(\boldsymbol{x};\theta)$, i.e.,
$p\left( \boldsymbol{X}_{N} \mid g(\boldsymbol{x};\theta) \right)=p\left( \boldsymbol{X}_{N}\right)$,
but the assumption breaks if there is selection bias \cite{gelman2014bayesian}.
Any choice on the parametric structure of the model $g(\boldsymbol{x};\theta)$, implicitly specifies $p\left(g(\boldsymbol{x};\theta)\right)$ in terms of the non-linear relationships among the covariates, smoothness, continuity etc.

For parametric models like linear regression where the structure of the model $g(\boldsymbol{x};\theta)$ is known, Eq. (\ref{eq:myeqdef1}) essentially transforms into an inference only on the model parameters $\theta$ given by
\begin{equation}
\label{eq:myeqdef2}
p\left( \theta \mid \boldsymbol{Y}_{N}, \boldsymbol{X}_{N} \right)
=
\frac{
       p\left( \boldsymbol{Y}_{N}\mid \theta, \boldsymbol{X}_{N} \right)
       p\left( \theta \right)
     }{
       p\left( \boldsymbol{Y}_{N} \mid \boldsymbol{X}_{N} \right)
      }
\end{equation}
For non-parametric approaches, like Gaussian processes, the prior 
$p\left( g(\boldsymbol{x};\theta) \right)$ can be placed directly on the function space without any requirement of explicit parameterisation of the function 
$g(\boldsymbol{x};\theta)$.
In our case, the function $g(\boldsymbol{x};\theta)$ is a fuzzy inference system given by the rule base Eq.  (\ref{eq:fuzzyrulebase})
\begin{equation}
\label{eq:fuzzyrulebase}
\beta_{k}R_{k}:if \:A^{k}_{1} \oplus A^{k}_{2} \oplus\dotso \oplus A^{k}_{T_{k}} \; then \; C_{k}
\end{equation}
where $\beta_{k}$ is a dichotomous variable indicating the inclusion of the $k^{th}$ rule in the fuzzy inference system, $A^{k}_{i} \;\forall i \in \left\{ 1,2,...,T_{k} \right\}$ is a referential value of the $i^{th}$ antecedent attribute in the $k^{th}$ rule, $T_{k}$ is the number of antecedent attributes used in the $k^{th}$ rule, $C_{k}$ is the consequent in the $k^{th}$ rule, $\oplus\in\left\{ \vee,\wedge \right\}$ represents the set of connectives (OR, AND operations) in the rules. A referential value is a member of a referential set with the universe of discourse $U_i$, which are meaningful representations for specifying an attribute using subjective linguistic terms which are easily understood by an expert. For example, an attribute representing maintenance level might consist of linguistic terms like `poor', `medium' and `good'. Each of these referential values are characterised by membership functions $f_{i}\left( A_{i},U_i, \phi_{i}\right) or f_{i}\left( C_{i}, U_i, \phi_{i}\right)$ and for the inference of the function $g(\boldsymbol{x};\theta)$ in Eq. (\ref{eq:myeqdef1}), the set of parameters to be estimated is 
$\theta$, which contains all $\phi_{i}$ and $\beta_{k}$

\begin{em} Remark 1: \end{em} Since the parameter estimation problem does not impose any specific structure on Eq. (\ref{eq:fuzzyrulebase}), the underlying fuzzy rule bases or sets can be defined to be of other types like fuzzy type II, intuitionistic fuzzy sets, hesitant fuzzy sets etc. 

\begin{em} Remark 2: \end{em} Including $\beta_{k}$  in Eq. (\ref{eq:fuzzyrulebase}) for selecting each rule and estimating these co-efficients within a Bayesian framework, essentially results in knowledge extraction from the data in the form of rules, i.e., the expert has some prior opinion on specific set of rules and the Bayesian inference engine combines the data with this to show which of the ones are more likely given the data.

\subsection{Regression on synthetic data-sets}

\subsubsection{Bayesian estimation of fuzzy membership function parameters}
Consider the synthetic example of predicting downtime of an engineering process system based on the variables location risk and maintenance level. For the  referential sets $loc\_risk,  maintenance,  downtime$ the referential values are $ \left\{ LO, MED, HI \right\}$ , $\left\{ POOR, AVG, GOOD\right\}$, $\left\{ LO, MED, HI\right\}$ respectively. The corresponding membership functions are considered to be triangular and can be mathematically expressed as 

\begin{equation}
\label{eq:triang_mem}
f_i\left ( u , \phi_{i}\right  )=\left\{\begin{matrix}
\frac{\left ( u-\phi_{i_1} \right )}{\left ( \phi_{i_2}-\phi_{i_1} \right )} \:\:  if \: \phi_{i_1}\leq u\leq \phi_{i_2} 
\\ 
\frac{\left ( u-\phi_{i_3} \right )}{\left ( \phi_{i_2}-\phi_{i_3} \right )} \:\:  if \: \phi_{i_2}\leq u\leq \phi_{i_3} 
\\
0 \: \: \: \: \: \: \: \: \: \: \: \:  otherwise
\end{matrix}\right.
\end{equation}

where, $ u \in U_i $ is any value within the universe of discourse and $ \phi_{i_j} \in \phi_{i} \forall j $ are the parameters of the membership functions to be estimated by the MCMC algorithm. The rule base used for constructing the fuzzy inference system is as follows:
\begin{equation}
\label{eq:rule123}
\begin{split}
R_{1}:if &\:loc\_risk==HI \; \vee \; maintenance==POOR \\
							  &then \; downtime==HI \\
R_{2}:if &\:loc\_risk==	MED \; \vee \; maintenance==AVG \\
							  &then \; downtime==MED\\
R_{3}:if &\:loc\_risk==LOW \; \wedge \; maintenance==GOOD \\
							  &then \; downtime==LOW\\		  
\end{split}
\end{equation}

For this case we assume all the rules to be known ($\beta_{k}=1, \forall k$) and only the parameters (\begin{math} \phi_{i} \end{math}) of the membership functions ($f_{i}$) need to be estimated. For this example, 100 data points are generated from the rule base in Eq. (\ref{eq:rule123}) using triangular membership functions and centroid method of de-fuzzification. The values of the tips and the bases of the triangular membership functions are to be estimated by the MCMC algorithm. Therefore, for this case with three referential sets each with three referential attributes, there are 9 parameters to be estimated.
Assume the following for the set of observations  $\{\boldsymbol{Y}_N,\boldsymbol{X}_N\}$:
\begin{equation}
\label{eq:normal:assumption}
\boldsymbol{Y}_N \sim  \mathcal{MVN} \left( g\left(\boldsymbol{X}_N;\theta \right) ,\sigma^2 I_N \right)
\text{,}
\end{equation}
where
$\mathcal{MVN} (\boldsymbol{\mu} , \Sigma)$
is the Multivariate-Normal distribution with mean vector $\mu$ and covariance matrix $\Sigma$.
The mean vector for our model comes from $g\left(\boldsymbol{X}_N;\theta \right)$ and the fuzzy inference system from Eq. (\ref{eq:fuzzyrulebase}).
We use $I_N$, the $N \times N$ identity matrix, along with a parameter $\sigma$ to construct the covariance matrix.
In other words, the observations of 100 data-points are assumed to come from a normal distribution with the mean ($\mu$) of the distribution given by the output of the fuzzy inference system and a variance ($\sigma$) representing the measurement error.
This leads to the likelihood
\begin{equation}
\label{eq:likelihood}
\begin{aligned}
&p\left( \boldsymbol{y}_{N}\mid g(\boldsymbol{x};\theta), \boldsymbol{X}_{N} \right)\\
&=
\frac{1}{\sqrt{(2\pi)^{N}\sigma^{2N} }} \times \\
&\phantom{=}
\exp\left(-\frac{1}{2\sigma^{2N} }
         \big(\boldsymbol{y}_{N}-g(\boldsymbol{X}_N;\theta)\big)^\mathrm{T}
         \big(\boldsymbol{y}_{N}-g(\boldsymbol{X}_N;\theta)\big)
\right)
\text{.}
\end{aligned}
\end{equation}
where $\boldsymbol{y}_N$ is a realisation of the random variable $\boldsymbol{Y}_N$.
The set of parameters is $\theta=\{\phi_i : i = 1 \ldots n\}$ where $n$ is the number of parameters to be estimated, and we can estimate it by being Bayesian and assuming a prior $p(\theta)$.
This gives the posterior distribution:
\begin{equation}
\label{eq:posterior}
p\left( \theta | \boldsymbol{y}_{N} ,  \boldsymbol{X}_{N} \right) \propto
p\left( \boldsymbol{y}_{N}\mid g(\boldsymbol{x};\theta), \boldsymbol{X}_{N} \right)p(\theta)
\text{.}
\end{equation}
Although Eq.~(\ref{eq:posterior}) is not mathematically tractable, we can draw samples from it using a Metropolis-Hastings algorithm such as the Gibbs sampler \cite{Robert2004}.

\paragraph{Case I: Parameter estimation without measurement noise and small data set}
\label{sec:case1}
For this case, since the data-points are generated from the true fuzzy inference system without any addition of noise, we set  $\sigma=0.001$, i.e. the measurement error is negligible.
We set the prior $p(\theta)$ to be independent uniform distributions over the parameter space for each $\phi_i$ 
\begin{equation}
 \phi_i \sim \text{Uniform}(0,C_{max})
\end{equation}
such that
\begin{equation}
\begin{aligned}
p(\theta) 
&= p(\phi_1,\phi_2, \ldots, \phi_n)  \\
&= p(\phi_1) p(\phi_2) \ldots p(\phi_n) \\
&= \left(\frac{1}{C_{max}}\right)^{n}
\text{.}
\end{aligned}
\end{equation}
Here, $C_{max}$ is the maximum allowed value for the parameter $\phi_i$, see table \ref{regression_hdi} for the values.

The first set of MCMC simulations are run with only 15 data points and the convergence diagnostics for one of the parameters (parameter 9) is shown below in Fig.~\ref{fig:15data}. The convergence curves for all the others parameters are similar and the Highest Density Intervals (HDIs) are tabulated in Table \ref{regression_hdi} in the appendix. For all the simulation cases, non-informative priors are used for all the parameters during the Bayesian estimation phase. For all the MCMC simulations in this paper, the python package pymc3 \cite{salvatier2016probabilistic} is used. 

\begin{figure}[htb!]
 \centering
 \includegraphics[width=0.99\columnwidth ]{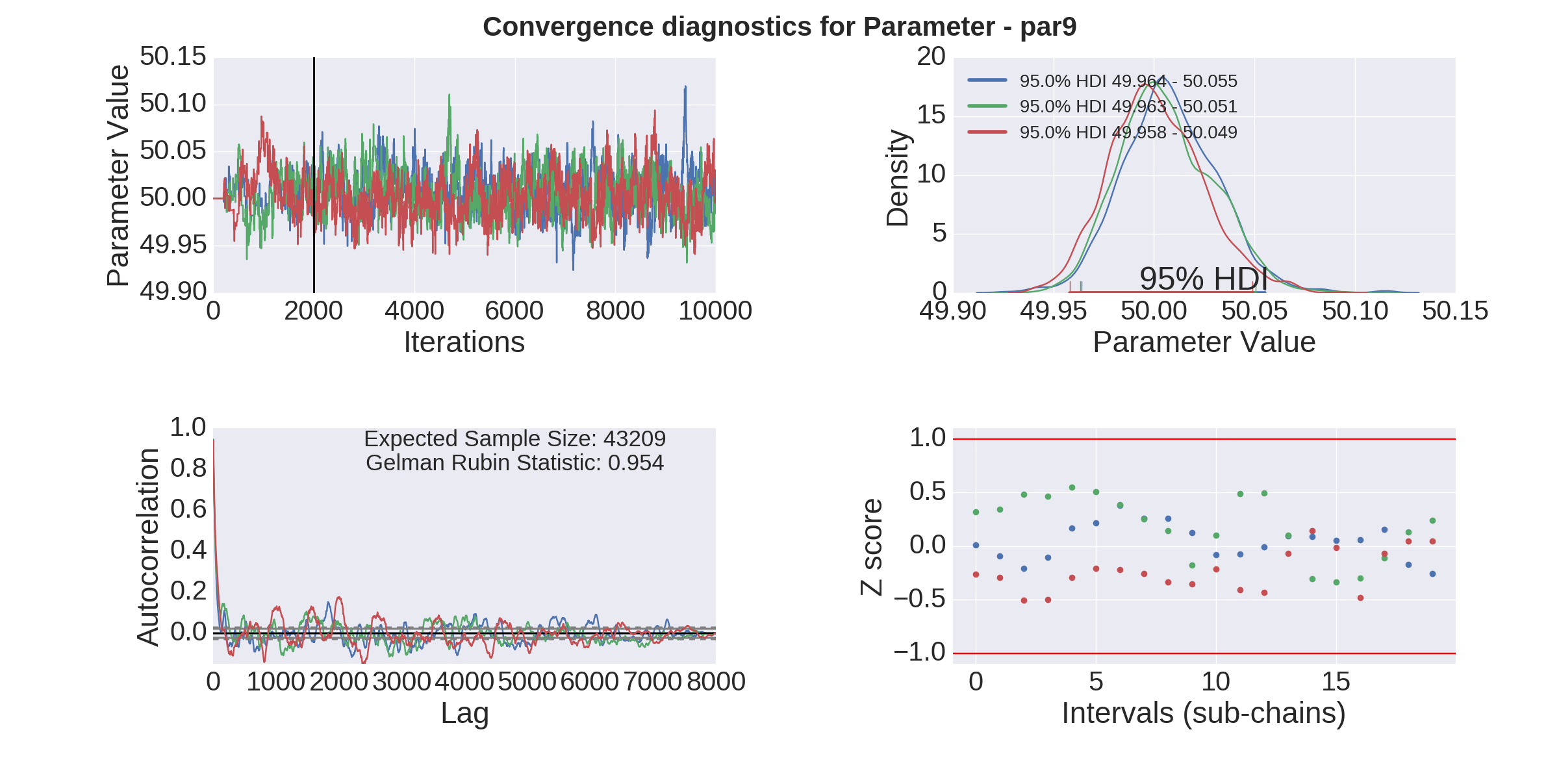}
 \caption{Case I: MCMC convergence diagnostics for parameter 9 with 15 data points}
 \label{fig:15data}
\end{figure}

The top left sub-plot in Fig.~\ref{fig:15data} shows the trace plots for three independent MCMC chains over the parameter space for 10,000 iterations. The burn-in is set at 2000 iterations and is indicated in the figure with a black line. The chains are well mixed and the density plots on the top right are overlapping and indicate that they are representative of the true posterior distribution. The 95\% highest density intervals (HDIs) are also plotted on the x-axis which represent the most probable range of the parameter value. For parameter 9, the true value was 50, and the HDIs for all the three chains are within 49.95 and 50.05, showing that the MCMC algorithm is able to correctly identify the true value of the parameter within a very tight range with only 15 data points. The convergence characteristics can also be checked with the auto-correlation plots of the three chains on the bottom left. The auto-correlation falls quickly with the number of lags for the post burn-in samples and therefore other methods like chain thinning are not required to obtain a good posterior distribution. The effective sample size (ESS) \cite{gelman2014bayesian} is also reported in this sub-plot which essentially reflects the sample size of a completely non-autocorrelated chain which would yield the same information as the one obtained from the MCMC algorithm. A higher ESS indicates better approximation of the posterior distribution.   

The Gelman-Rubin statistic or the shrink factor is a measure of the variance between the chains relative to the variance within the chains. The value for this case, as indicated in the figure, is near 1 which implies that the chains have converged. As a thumb rule, if the value is more than 1.1, then there are orphaned or stuck chains.  
The Z-scores given in the convergence diagnostic plots are the Geweke's diagnostic.
We expect these to be from a normal distribution, so most values should be within the red lines in the range [-1,1].
If a substantial amount of the Z-scores fall outside of this, it implies a failure in convergence.

\paragraph{Case II: Parameter estimation without measurement noise and more data points}

\begin{figure}[h!]
 \centering
 \includegraphics[width=0.99\columnwidth ]{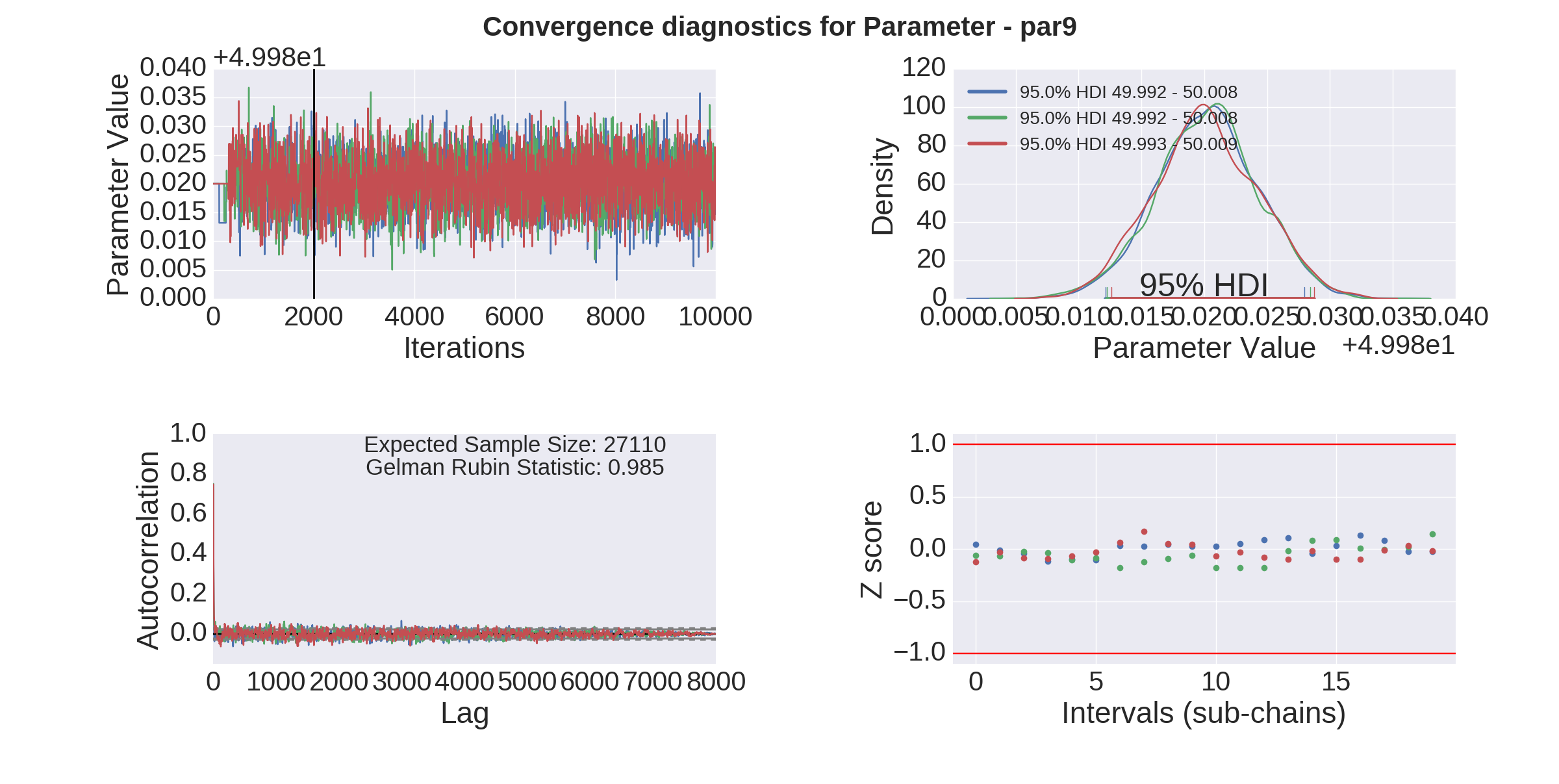}
 \caption{Case II: MCMC convergence diagnostics for parameter 9 with 100 data points}
 \label{fig:1000data}
\end{figure}

The next set of simulations are done with the same settings as the previous case but with 100 data-points instead of 15. The convergence characteristics of the same parameter (par9) is shown in Fig.~\ref{fig:1000data} and as can be seen in the top right density sub-plot, the 95\% HDI is very small around the true value of 50, demonstrating that more data-points help in obtaining much tighter estimates of the parameter values.
The trace plots, auto-correlation plots and the Geweke scores indicate that the simulations have converged.

\begin{figure}[h!]
 \centering
 \includegraphics[width=0.99\columnwidth ]{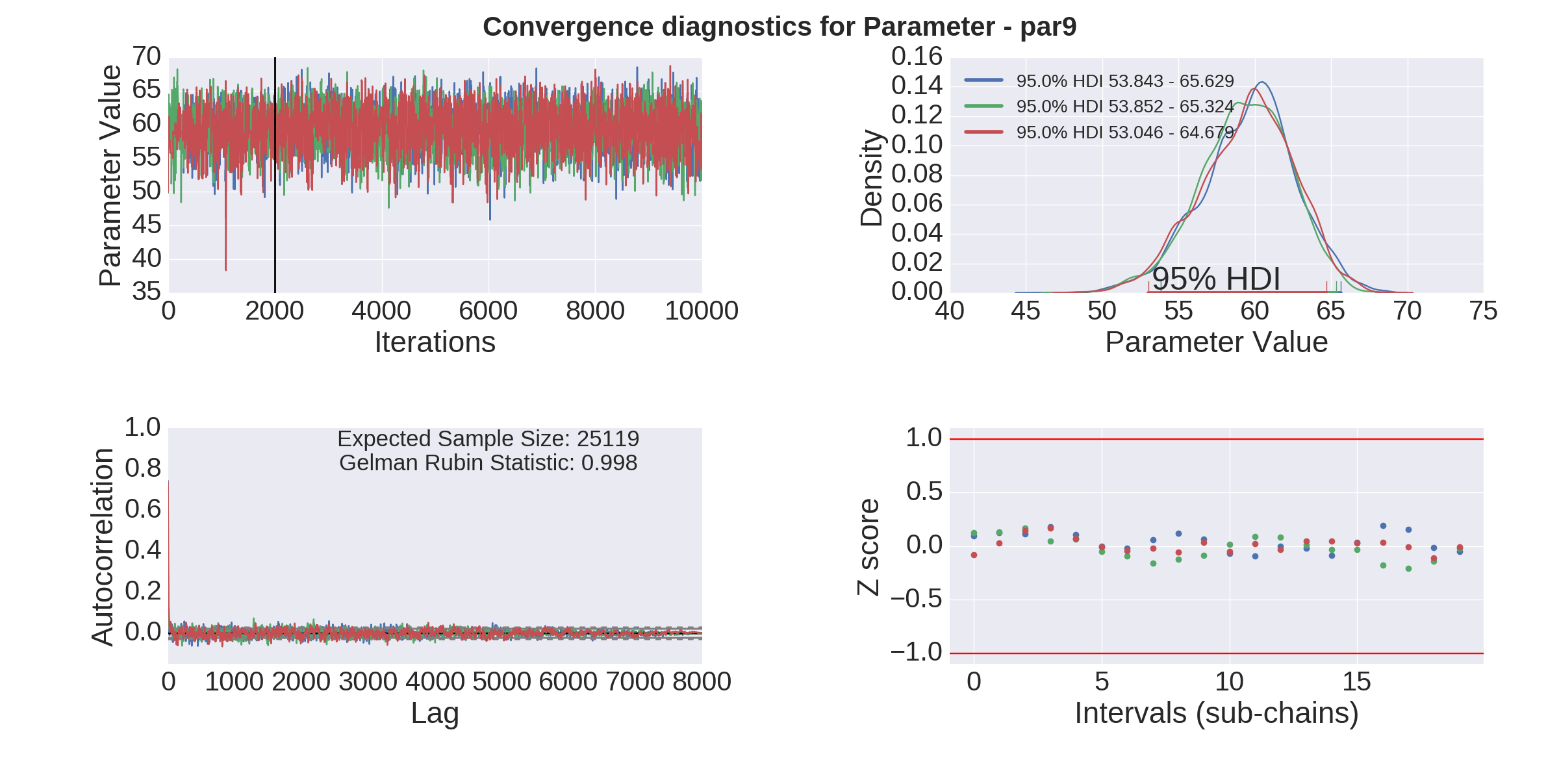}
 
 \caption{Case III: MCMC convergence diagnostics for parameter 9 with 100 data points and added noise}
 \label{fig:noise1000data}
\end{figure}

\paragraph{Case III (a): Parameter estimation with additional measurement noise}
For this case, noise from a distribution of $\mathcal{N} \left( 0,1 \right)$ is added to the 100 observations. The likelihood function for the MCMC in Eq. (\ref{eq:normal:assumption}) then has $\sigma=1 $.
The convergence diagnostics with the density plots for the same parameter (par9) is shown in Fig. \ref{fig:noise1000data}.

\begin{figure}[h!]
 \centering
 \includegraphics[width=0.99\columnwidth ]{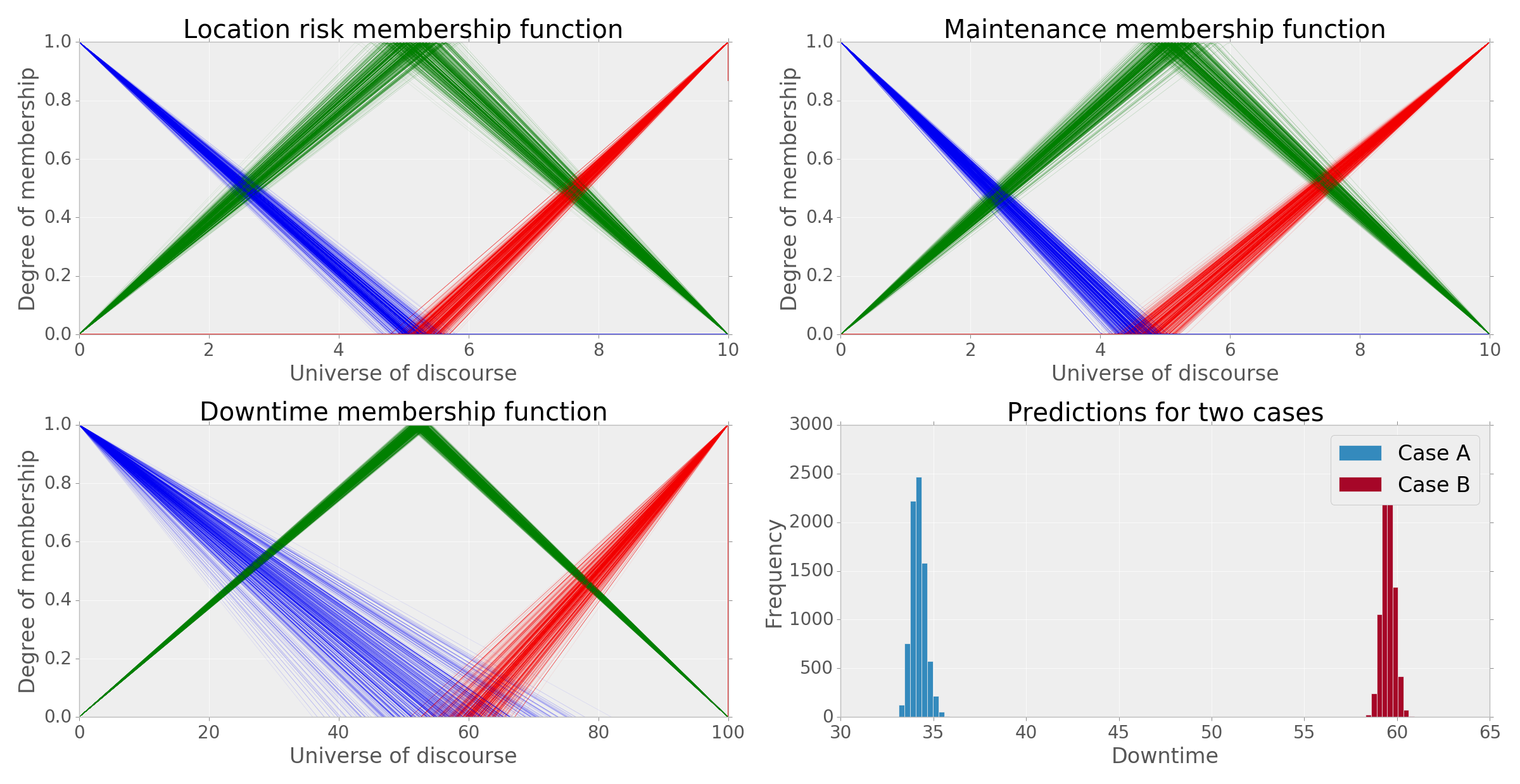}
 \caption{Posterior of the membership functions and predictive distribution for two cases A and B. Fuzzy sets with blue colour indicate LO or POOR, green colour indicate MED or AVG and red colour indicate HI or GOOD.}
 \label{fig:memfunc}
\end{figure}

It can be seen that the HDI of the parameter estimate now spans a much larger region than the two previous simulation runs, due to the noisy data-points. The uncertainty in parameter estimates can be seen from the plots of the membership functions obtained by plugging in the values from the posterior distribution, as depicted in Fig.~\ref{fig:memfunc}.
It can be observed that for some of the membership function values like the LOW and HI membership function for Downtime, the uncertainties are much higher than the other ones. This might indicate that either there were not enough samples in that region to have more tighter estimates for these membership functions, or due to the non-linear nature of the fuzzy rule base, the output variable (Downtime) and its membership functions have more flexibility and can map the final output for a wider range of parameters.   

The bottom right sub-plot in Fig.~\ref{fig:memfunc} shows the predictions of the Downtime for two different scenarios (Cases A and B). For Case A, the input variables are set to [1.1,8.8] for location risk and maintenance respectively. For Case B, the values are set to [7.7,1.1]. Therefore, in Case A the input suggests that the location risk is low while the maintenance is good. Hence we would intuitively expect the downtime to be low. For case B, the values suggest that the location risk is high and the maintenance is poor. Hence we would intuitively expect the Downtime to be high. This is confirmed by looking at the predictions obtained from the posterior samples as shown in the histograms at the bottom right sub-plot in Fig.~\ref{fig:memfunc}. For Case A, the Downtime is low with a distribution around 34 and for Case B, the Downtime is high with a distribution around 59.

Once the parameter estimates are obtained, the prediction over the whole universe of discourse for all the variables can be obtained along with the posterior distribution of the uncertainty at each point as shown in Fig.~\ref{fig:minmaxsurf} in the supplementary materials.

\paragraph{Case III (b): Parameter estimation with additional measurement noise, when the noise parameter is unknown}
This case is similar to Case III (a), except that we assume $\sigma$, the noise parameter, is unknown and has to be estimated along with the rest of the parameters.
In the estimation, a vague prior ($\sigma \sim \text{Uniform}(0.01, 10)$) is used.
The results show that the MCMC algorithm is able to identify the true value of $\sigma$ to reasonably good accuracy with tight HDI bounds.
The parameters of the fuzzy membership functions themselves are also identified, though the uncertainty bounds on them (as measured through HDI) are a bit larger than Case III (a), where the value of $\sigma $ was specified to the true value.
This matches with our intuitive understanding of the problem; adding more parameters while keeping the amount of data constant leads to more uncertainty in the estimates.
Thus, the FBL technique is not overly sensitive to the value of $\sigma$ in the present simulation.
Care has to be taken with $\sigma$'s prior.
If it is completely erroneous (e.g. range of prior does not include the true value etc.), then the algorithm would fail (or give completely erroneous results); as is the case for any Bayesian analysis with wrong priors.
In cases where the signal to noise ratio is too low, estimating the value of $\sigma$ would be difficult and the analysis would not provide useful outputs.

\subsubsection{Bayesian rule base selection and parameter estimation of membership function (Case IV)}
\label{subsubsec:rule}
Consider the case where the dataset is generated as before from the three rules in Eq. (\ref{eq:rule123}), but two additional spurious rules are considered while estimating the parameters using MCMC. They are given by: 
\begin{equation}
\label{eq:rule45}
\begin{split}
R_{4}:if &\:loc\_risk==LO \;  \\
                &then \; downtime==HI \\
R_{5}:if &\:maintenance==POOR \\
                &then \; downtime==LOW\\
\end{split}
\end{equation}

Clearly these two rules are contrary to our intuition of maintenance and breakdown events mapped on to system downtime and are antagonistic to the original three rules given by Eq. (\ref{eq:rule123}). For this case, we set up the MCMC to not only select the parameters of the membership functions ($\phi_i$) but also the $\beta_k$ in Eq.~(\ref{eq:fuzzyrulebase}) which indicates the specific rules to be included for best explaining the dataset. 
Following the lines of \cite{RJMCMC}, we set up a reversible jump MCMC by assuming a prior on the $\beta_k$ parameters, which controls the inclusion of rule $k$.
We assume that the inclusion of any rule is independent of the inclusion of other rules, and that rules are equally likely to be included as excluded.
Probabilistically, this is represented as:
\begin{equation}
\beta_k \sim \text{Bernoulli}(0.5) 
\end{equation}
and the $\beta_k$ parameters can be sampled from the posterior distribution along with the other parameters.

\begin{figure}[h!]
 \centering
 \includegraphics[width=0.99\columnwidth ]{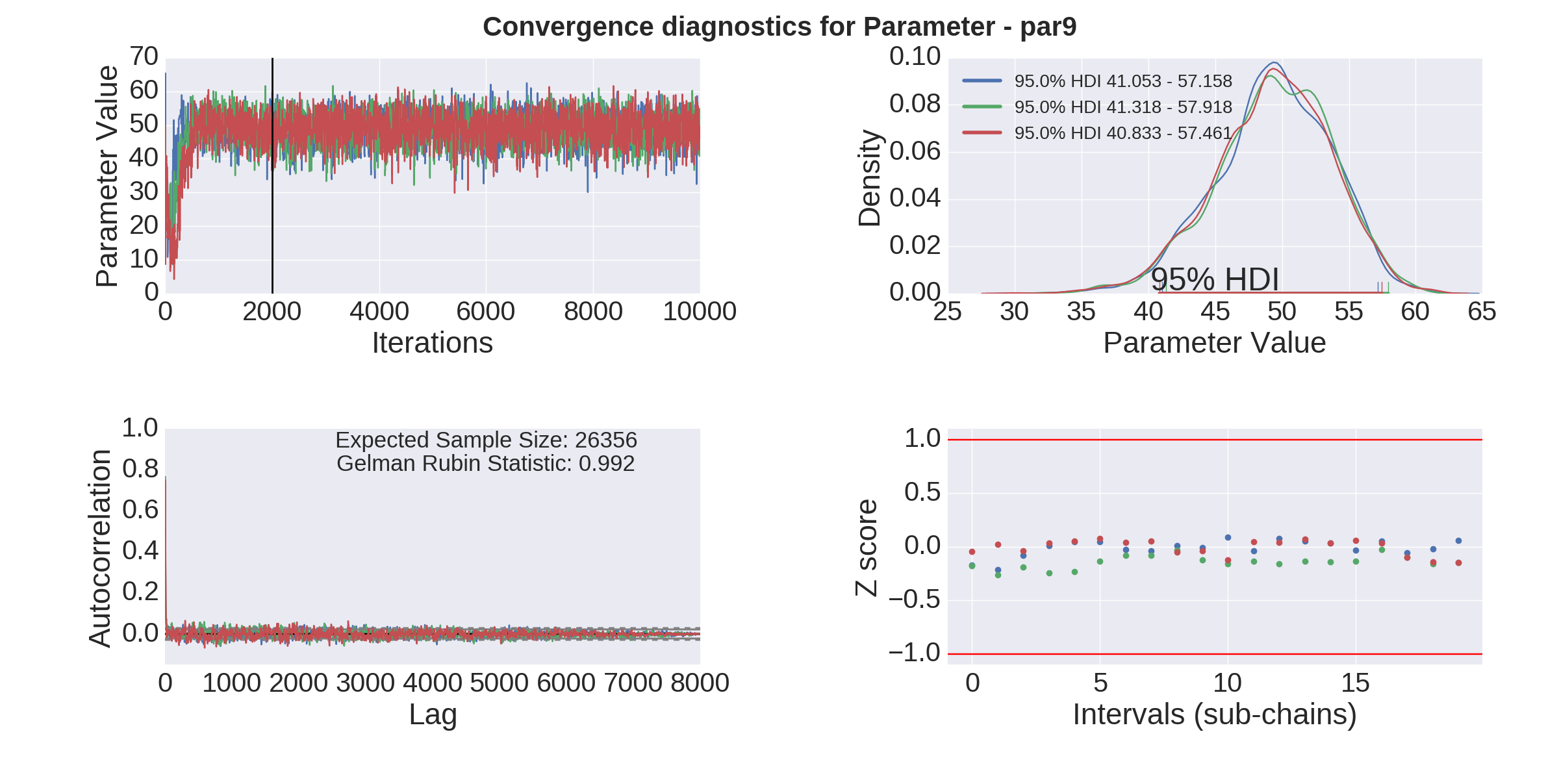}
 \caption{Case IV: MCMC convergence diagnostics of parameter 9 with for the case of rule base selection}
 \label{fig:ruleselecpar9}
\end{figure}

As before, we ran the MCMC algorithm for 10,000 iterations and the convergence diagnostics of the same parameter (par9) is shown in Fig.~\ref{fig:ruleselecpar9}.  The effect of the burn-in is clear from the top left plot and the parameter estimates in the top right have a higher variance (compared to the previous cases) around the true value of 50. This is due to the fact that the same 100 data points are used to estimate the parameters as before, but in this case the number of parameters are greater -- 14 parameters (9 parameters for the fuzzy membership functions and 5 parameters for selecting the individual rules) as opposed to just 9 parameters in the previous cases. The algorithm tries to extract more information from the data (more parameters to estimate) and hence the uncertainties in the parameters are higher. More data would decrease uncertainty around the estimates.  

\begin{figure}[h!]
 \centering
 \includegraphics[width=0.99\columnwidth ]{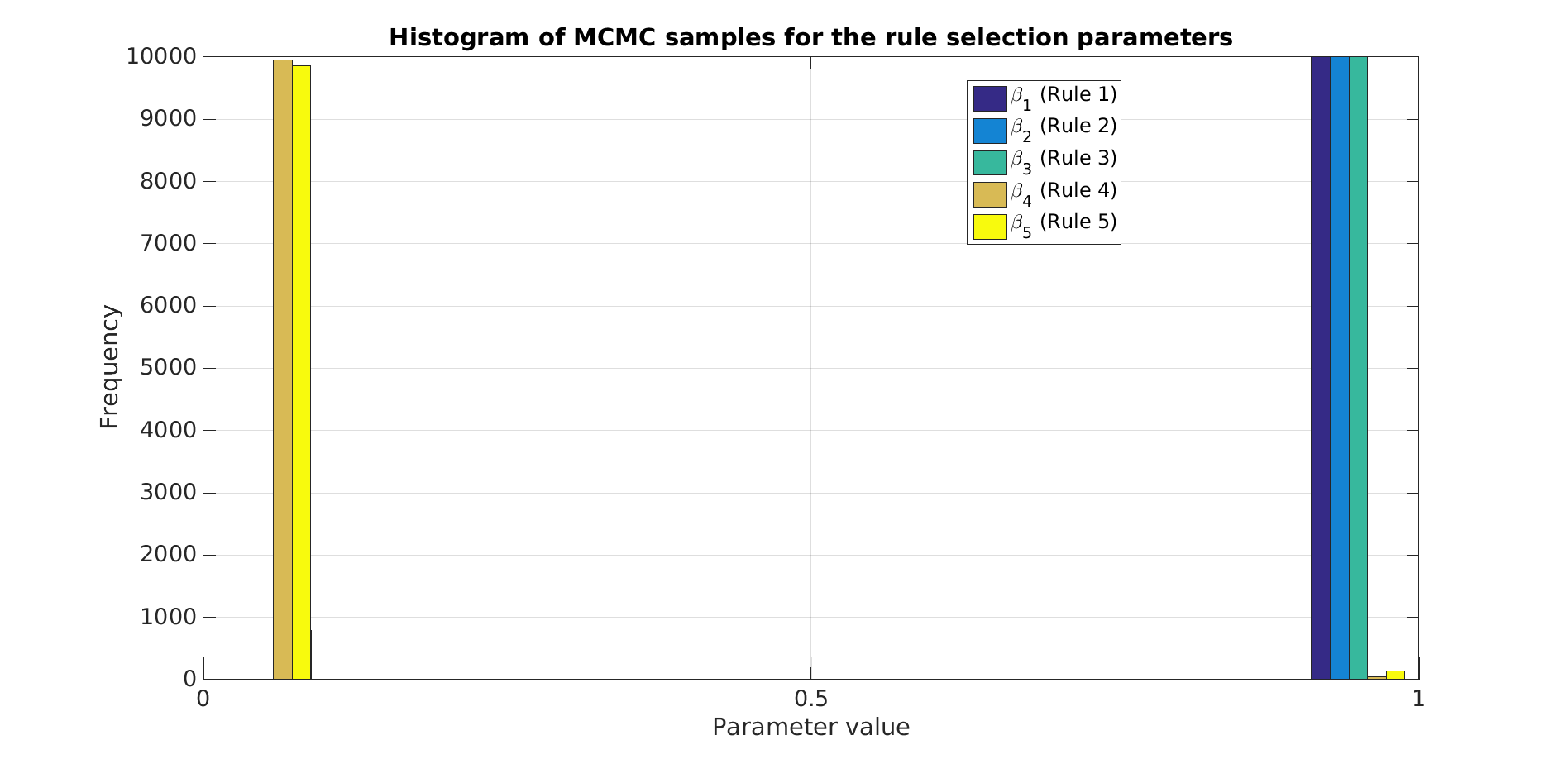}
 \caption{Histogram of the rule base selection parameters. All values are binary (0 or 1), and due to the stacked nature of the plot they appear on the continuum of the abscissa.}
 \label{fig:ruleselechist}
\end{figure}

Since the rule selection parameter is a Bernoulli distribution (essentially 0 or 1 values instead of continuous real values), it is easier to interpret the results from a histogram plot (Fig.~\ref{fig:ruleselechist}) of the MCMC samples for these parameters. 

Fig.~\ref{fig:ruleselechist} shows that the two spurious rules (Rule 4 \& 5) were not chosen by the algorithm (most of the MCMC samples for these have zero values) and the other three rules which generated the data (Rule 1, 2, 3) were correctly identified by the algorithm (most values are one). A small number of MCMC samples for Rules 4, 5 have values of one as well, which is during the initial exploration phase of the MCMC (the plot shows values for all the samples and not just post burn-in) and they quickly converge to the true values. 

\subsubsection{FBL for the case of sparse rule bases}
Next we show how the FBL can work for inference systems with sparse membership sets and with missing rules like the canonical example of the tomato classification problem as discussed in \cite{mizumotocomparison, hsiaonew}. The problem with such cases is that the rule bases are incomplete and there are empty spaces between membership functions of the antecedants so that when the membership function of the observation falls in the empty space, no rule is fired and therefore no output or consequent is obtained \cite{yanreasoning, hsiaonew}. Following the example of \cite{mizumotocomparison, hsiaonew} we define the  referential sets $colour, ripeness$ the referential values $ \left\{ GREEN, YELLOW, RED \right\}$ , $\left\{ UNRIPE, HALF-RIPE, RIPE \right\}$, respectively, using triangular membership functions as in Eq.~(\ref{eq:triang_mem}), without any overlap. The full rule base is given as Eq.~(\ref{eq:rule_tomato}), which includes all the three rules $\left\{R_{a},R_{b},R_{c}\right\}$ from which 100 data points are generated (which serves as the observations) and the MCMC algorithm is run on this rule base and data-set.

\begin{equation}
\label{eq:rule_tomato}
\begin{split}
R_{a}:if &\:colour== GREEN \;  \\
                &then \; ripeness==UNRIPE\\
R_{b}:if &\:colour== YELLOW \;  \\
                &then \; ripeness==HALF-RIPE\\
R_{c}:if &\:colour== RED \;  \\
                &then \; ripeness==RIPE\\
\end{split}
\end{equation}

\begin{figure}[h!]
 \centering
 \includegraphics[width=0.99\columnwidth ]{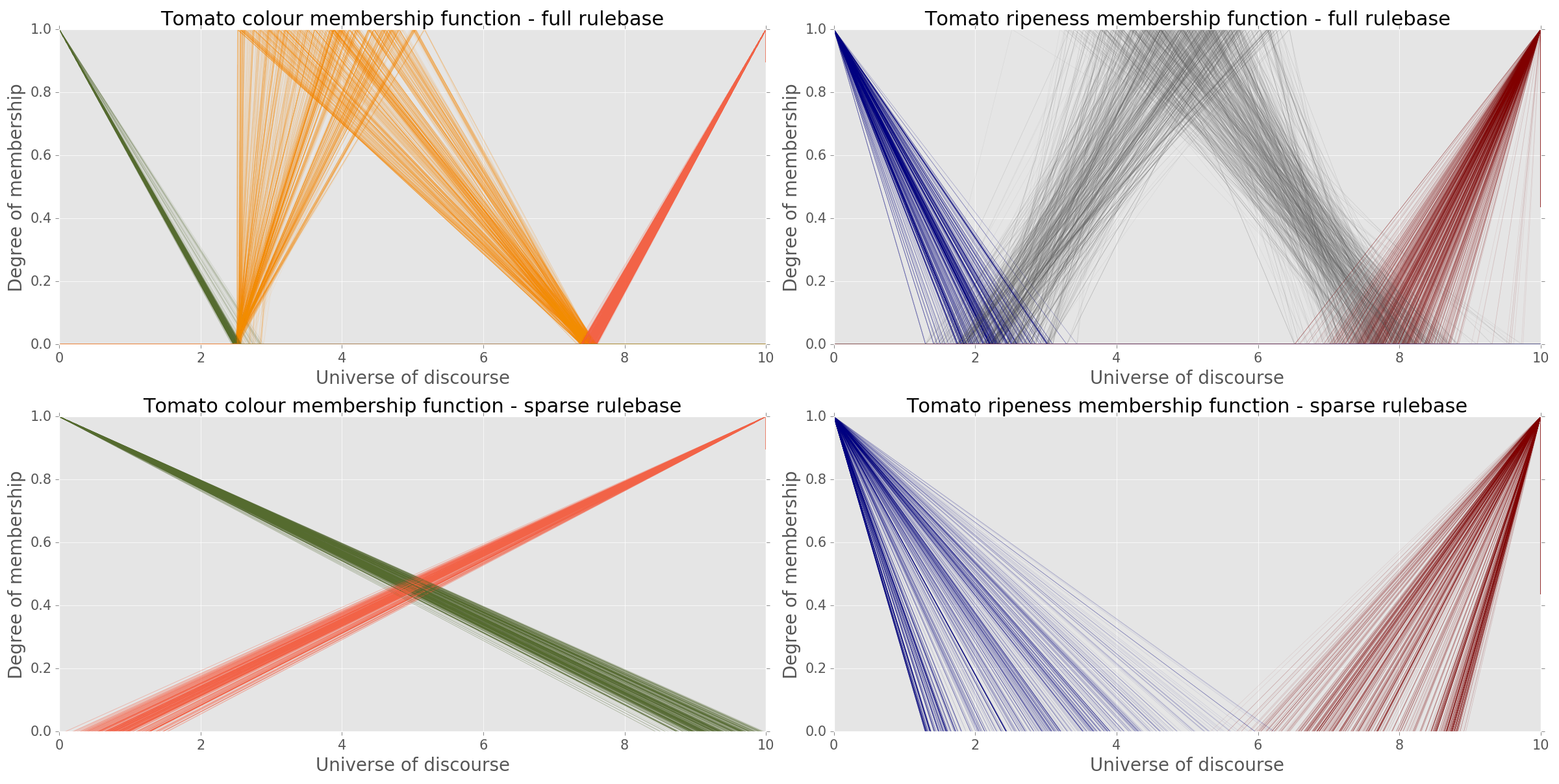}
 \caption{Posterior of the membership functions for the sparse rule base case. Colour coding for membership functions: green, golden and orange colours indicate the referential values GREEN, YELLOW and RED respectively; blue, gray and maroon indicate UNRIPE, HALF-RIPE and RIPE respectively.}
 \label{fig:sparse_rules}
\end{figure}

The results (posterior of the membership functions) are shown in the top row of Fig.~(\ref{fig:sparse_rules}) which depicts that the true underlying membership functions can be approximated with the full rule base. The second set of simulations is run with only two rules $\left\{R_{a},R_{c}\right\}$ and the referential values of $YELLOW, HALF-RIPE$ are deleted from the referential sets $colour, ripeness$. The results from the MCMC simulations using the previous 100 data points are shown in the bottom row of Fig. (\ref{fig:sparse_rules}). As can be observed, the empty space for the value $YELLOW$ is made up for in the antecedant by significant overlap of the other two sets. Also the empty space for $HALF-RIPE$ is taken care of by the other two sets in the consequent which are much wider now to cover the whole space. Therefore we can see that in the case of sparse rules and membership functions, the MCMC algorithm automatically adjusts the membership functions in such a way that they cover the empty spaces and does a sort of smart interpolation to best explain the data with the sparse rule-base.

Of course, more complicated interpolation rules can be incorporated in FBL if it is desired to have, for example, a triangular-type interpolated conclusion for triangular-type observations as proposed in \cite{hsiaonew}. The framework proposed here-in is generic to accommodate for such intricacies albeit at the cost of higher model complexity.

\subsubsection{Algorithmic details of MCMC for FBL and its estimation properties}

We present a brief discussion on the computational details of the MCMC algorithm, given in algorithm \ref{algo:MCMC}.
\begin{algorithm}
\caption{MCMC Gibbs Sampling}
\label{algo:MCMC}
\begin{algorithmic}[1]
\renewcommand{\algorithmicrequire}{\textbf{Input:}}
\renewcommand{\algorithmicensure}{\textbf{Output:}}
 \Require $G$ (Number of samples required),
 \Ensure $G$, Posterior samples of $\boldsymbol{\theta}$
\State $\boldsymbol{\theta} \gets$ arbitrary starting values
\Comment Initialisation

\For{$G$ iterations}
\For{$i=1$ \textbf{to} $n$}
\Comment $n$ is number of parameters in $\boldsymbol{\theta}$
  \State $\theta_i \gets$ DrawSample( $p\left(\theta_i| \boldsymbol{\theta}_{/\theta_i} \right), \theta_i$ )
  \State Store($\theta_i $)
  \Comment Write to MCMC chain
\EndFor
\EndFor

\end{algorithmic}
\end{algorithm}
Here, DrawSample() is a function that takes a distribution and an optional current value of $\theta_i$, and
gives a new sample of $\theta_i$ from the given distribution.
The distribution
$p\left(\theta_i| \boldsymbol{\theta}_{/\theta_i} \right)$
is the conditional distribution of $\theta_i$ given all other parameters in $\boldsymbol\theta$, excluding $\theta_i$, where $n$ is the number of parameters in the model:\\

\[
p\left(\theta_i| \boldsymbol{\theta}_{/\theta_i} \right) =
    p\left(\theta_i
            \big|
            {\theta}_1,
            \ldots,
            {\theta}_{i-1},
            {\theta}_{i+1},
            \ldots,
            {\theta}_{n  }
      \right)
\text.
\]
This will be proportional to the joint distribution:
\begin{align*}
p\left(\theta_i| \boldsymbol{\theta}_{/\theta_i} \right)
&=
\frac{
      p\left({\theta}_1, \ldots, {\theta}_{n}  \right)
     }
     {
      p\left({\theta}_1,
             \ldots,
             {\theta}_{i-1},
             {\theta}_{i+1},
             \ldots,
             {\theta}_{n}
        \right)
     }
\\
&\propto
p\left({\theta}_1, \ldots, {\theta}_{n}  \right)
\text.
\end{align*}
The proportionality is important, since the DrawSample() function is usually the Metropolis-Hastings algorithm (or one of its derivates), which only requires the target density to a normalising constant.
In our case, this means we have to pass the density in
Eq.\ (\ref{eq:posterior}) to DrawSample(), which includes $g(\boldsymbol{x}; \boldsymbol{\theta})$.
The DrawSample() function usually involves proposing new values, and accepting/rejecting these values, as shown in algorithm \ref{algo:drawsample}.
\begin{algorithm}
\caption{Sample from arbitrary density function}
\label{algo:drawsample}
\begin{algorithmic}[1]
\Procedure{DrawSample}{$p(\cdot)$, $\theta_{\text{current}}$}
\For{$L$ iterations}
\Comment Some limit to avoid infinite loop
  \State Propose new sample $\theta_{\text{new}}$ using $\theta_{\text{current}}$
  \State Calculate acceptance probability using
  $p(\theta_{\text{new}})$ and
  $p(\theta_{\text{current}})$
  \If{New sample accepted}
   \State \Return{$\theta_{\text{new}}$}
  \EndIf
\EndFor
   \State \Return{$\theta_{\text{current}}$}
\EndProcedure
\end{algorithmic}
\end{algorithm}
Candidates for the DrawSample() function include 
Metropolis-Hastings and Slice sampling.
When comparing the GLM to the fuzzy model, the only difference is the specification of
$g(\boldsymbol{x}; \boldsymbol{\theta})$.
The GLM has a simple function, defined in terms of a link function and a linear equation and is computationally inexpensive since the predictions can be made using efficient matrix multiplications.
The fuzzy model has to re-compile each time the rule-base or its parameters changes, which is more computationally expensive.

\begin{figure}[h!]
 \centering
 \includegraphics[width=0.99\columnwidth ]{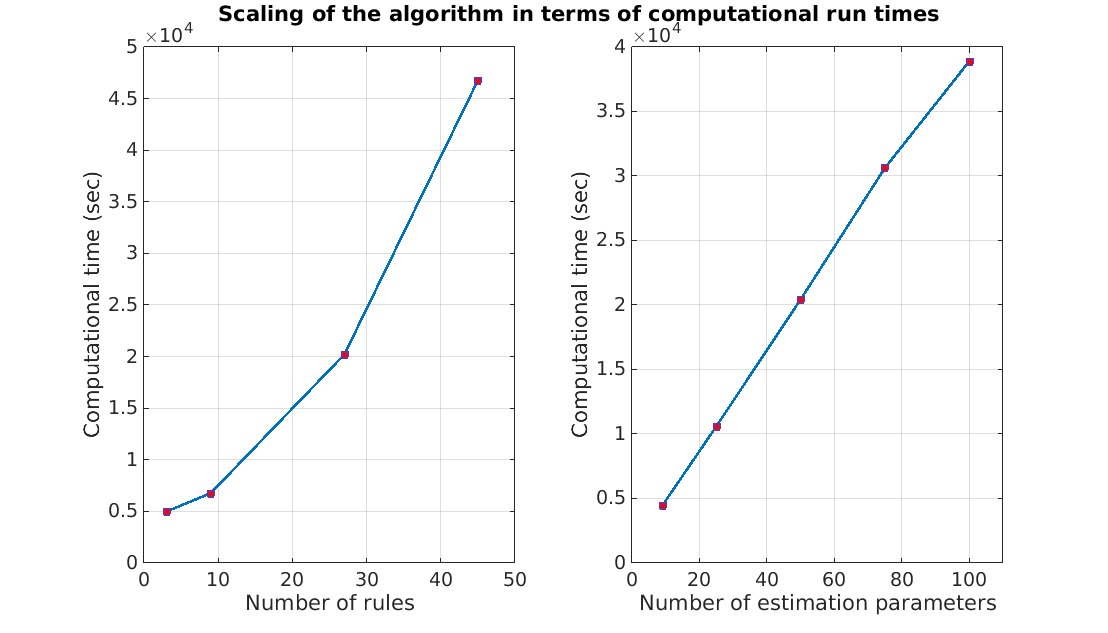}
 \caption{Comparison of computation times with respect to number of estimation parameters and rule base complexity}
 \label{fig:comp_time}
\end{figure}

To demonstrate the scaling of the MCMC algorithm with FBL in terms of computational time, we construct two different cases - (i) varying the number of estimation parameters and (ii) varying number of rules in the rule base. For case (i), the 15 data points are generated using the rules in Eq. (\ref{eq:rule123}) as in Case I of section (\ref{sec:case1}). Dummy estimation variables are added in the MCMC code so that the problem dimension increases but only nine of the original variables actually contribute to the fuzzy function. For case (ii), dummy input variables are added to the fuzzy inference system with fixed membership functions  with dummy rules, so that the complexity of the rule base increases, but number of estimation parameters remain the same. MCMC simulations are performed for 5000 iterations for each case on a 2.40 GHz Intel Xeon CPU with 16 cores and 24 GB RAM running Ubuntu 14.04 Server. The mean of three independent runs are reported in Fig. (\ref{fig:comp_time}) which shows that the computational time increases in a near exponential fashion with the number of rules in the fuzzy inference system while the increase is almost linear with respect to the number of parameters to be estimated by MCMC. This implies that there is more room for improvement of overall computational time by speeding up the fuzzy function evaluation. 

In our synthetic example (Case I in Sec. \ref{sec:case1}), we knew the true parameter values, thus we used a simulation study with 30 generated datasets to check if we were consistently over or underestimating the true parameters, using the posterior probability 
$p(\theta < \theta_{true} | \{\boldsymbol{Y}_N,\boldsymbol{X}_N\})$ calculated from the MCMC chains for each parameter.
With no systematic bias, these probabilities should adhere to a $\text{Uniform}(0,1)$ distribution.
We used a Chi-squared test on our posterior probability estimates, and it did not reject the null hypothesis of a uniform distribution, showing that the estimates do not have a systematic bias.

\subsubsection{Comparison of the method with Generalised Linear Models (GLM) using Bayesian parameter estimation}
\begin{figure}[h!]
 \centering
 \includegraphics[width=0.99\columnwidth ]{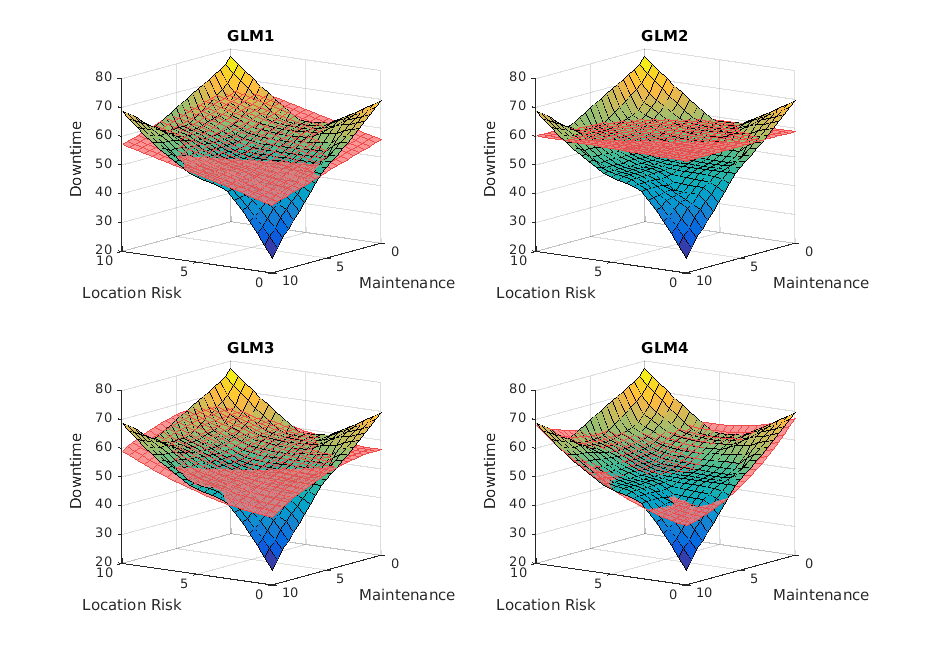}
 \caption{Comparison of the mean surface of the predictions (red surface) from GLM posteriors with the interpolated surface of the actual data-points}
 \label{fig:GLMsurf}
\end{figure}

To show the effectiveness of the proposed method, we compare four GLM structures with Bayesian parameter estimation. 
A GLM is composed of the following three elements: an error distribution, a link function, and a linear predictor \cite{GLMbook}.
We use the normal distribution as our error distribution:
\begin{equation}
\label{eq:GLMgeneric}
Y_i \sim  \mathcal{N} \left( \mu ,\sigma^2 \right)
\end{equation}
We use the identity link function to link the expected value to the linear predictor.
We test four GLMs, with linear predictors as given by the following equations, where $\alpha_i \forall i$ are the parameters to be estimated using MCMC.
\begin{equation}
\label{eq:GLM1}
GLM1: \mu=\alpha_0 + \alpha_1 x_1 + \alpha_2 x_2
\end{equation}
\begin{equation}
\label{eq:GLM2}
GLM2: \mu=\alpha_0 + \alpha_1 x_1 x_2
\end{equation}
\begin{equation}
\label{eq:GLM3}
GLM3: \mu=\alpha_0 + \alpha_1 x_1^2 + \alpha_2 x_2^2 + \alpha_3 x_1 x_2
\end{equation}
\begin{equation}
\label{eq:GLM4}
GLM4: \mu=\alpha_0 + \alpha_1 x_1 + \alpha_2 x_2 + \alpha_3 x_1^2 + \alpha_4 x_2^2 + \alpha_5 x_1 x_2
\end{equation}
For the MCMC sampling, the priors on the parameters were $\sigma \sim \text{half-Cauchy}\left(10 \right)$  and $\alpha_i \sim  \mathcal{N} \left( 0 ,20 \right) \forall i$, which are noninformative priors, ensuring that most of the information would come from the data.
The half-Cauchy distribution is defined as the Cauchy distribution left-truncated at zero, as given in \cite{gelman2006halfcauchy}.
The number of MCMC iterations are set to 2,000 and the burn-in as 500. The results are checked for convergence as before and the posterior samples after burn-in are plugged in to each of the GLMs and a distribution of the Downtime predictions over the whole 2D surface of Location Risk and Maintenance is obtained. The mean of these surfaces is plotted superimposed with the interpolated surface of the data points as shown in Fig.~\ref{fig:GLMsurf}.

As can be observed from Fig.~\ref{fig:GLMsurf}, as more interactions and parameters are added in the GLM structure, it has more flexibility in fitting the data and consequently the fits are better, with GLM4 giving the closest fit among all others. However, as can be seen from the actual structure of GLM4 in Eq.~\ref{eq:GLM4}, not much can be gleaned to relate to the expert opinions as expressed in Eq.~\ref{eq:rule123} from which the data was generated. In essence, the GLM is one of the many proxies that can fit the data.  
Also since the surface plots are interpolated from the data-points themselves, they are an approximation of the real surface which generated the data and would be more closer to the real one as and when more data-points are gathered. This might also imply that as more data points are gathered, a different GLM structure might fit better. This can also happen with the fuzzy rules where the algorithm has to choose the rules (as shown in the previous section), but if the algorithm selects a different fuzzy rule on having more data points, there is a more transparent explanation of the dynamics in the model.

In short, it is difficult to come up with a GLM model structure to explain such non-linear relationship in the data. In general, modellers experiment with a combination of linear models, models with interactions, models with higher order terms etc. The only way to include expert opinion in a traditional GLM is to use priors on the parameters (which is a lot less flexible that the IF-ELSE structured fuzzy rule base method proposed in the paper). But since the structure of a GLM is decided ad-hoc it is unclear how to transform expert opinions into priors or model parameters.

\subsubsection{Comparison with other popular frequentist machine learning methods}
\label{sec:freq_comparison}

For the sake of comprehensiveness, we also compare the FBL method with some of the popular methods in machine learning which are used in a similar way to learn from data and the corresponding results are reported in Table \ref{table_frequentist_comparison}. We use the small dataset (15 data points) for training as in Case I of \ref{sec:case1} and use a test dataset of 5 data points. The Adaptive Neuro Fuzzy Inference System (ANFIS) is trained with 3 membership functions for each of the inputs with 30 epochs. The neural network is trained with a Levenberg-Marquardt algorithm, has a single hidden layer with 10 neurons and 2 neurons in cases A and B respectively. The support vector regressor (SVR) is trained with a radial basis function kernel. \\
The key issue is that all of the other methods give point estimates and not probabilistic predictive distributions (like the FBL). Therefore the mean of the predictive distribution from the FBL is used to calculate the MSE (mean squared error) of prediction to compare the performance with other methods. There is no way to compare the uncertainties of the predictions. The results in Table \ref{table_frequentist_comparison} show that in most of the methods the training MSEs are generally quite low (due to the small dataset), but the test MSEs are much higher (i.e. methods do not generalise well to unseen data). The neural net A case is much worse than neural net B in test MSE due to higher number of neurons (it actually has more parameters than data points which would result in overfitting). None of the neural net or SVR gives rule bases like FBL or ANFIS. The ANFIS rule bases have the additional problem that they are generated from the data and are not always intuitive. The FBL method gives good train and test MSE since the rule base is correctly specified, which might not always be the case in real world cases as discussed later in Section \ref{potential_problems}.

\begin{table}[!th]
\renewcommand{\arraystretch}{1.3}
\caption{Comparison of mean squared prediction errors for the Fuzzy Bayesian model with other machine learning methods}
\label{table_frequentist_comparison}
\centering
\begin{tabular}{|c|c|c|c|}
\hline
Model & Prediction Type & Train MSE & Test MSE \\
\hline
ANFIS & Point predictions & $3.84 \times 10^{-9}$ & 7.857 \\
\hline
SVR & Point predictions & $2.31 \times 10^{-2}$ & $1.22$\\
\hline
Neural Net A & Point predictions & $7.31 \times 10^{-2}$ & 10.378\\
\hline
Neural Net B & Point predictions & $5.38 \times 10^{-1}$ & 0.810\\
\hline
FBL & Predictive distribution & $5.93 \times 10^{-5}$ & $7.64 \times 10^{-5}$ \\
\hline

\end{tabular}
\end{table}

\subsection{Classification on synthetic data-sets}

Our modelling methodology for regression can be easily extended for classification tasks. This extension is quite similar to using a logistic regression as a binary classifier. The binary classification task can be defined in terms of a Bernoulli likelihood function whose parameter ($\psi$) is set by the fuzzy inference system ($g\left( \cdot \right)$) as shown in Eq. \ref{eq:classif}.

\begin{equation}
\label{eq:classif}
\begin{split}
\boldsymbol{Y}_N \sim  \text{Bernoulli}\left( \psi \right) \\
\psi=g\left(\boldsymbol{X}_N;\theta \right) \\
\end{split}
\end{equation}
The rationale for this model for the classification task, is that the Bernoulli distribution gives either a one or a zero value (essentially representing the output class) depending on the probability parameter $\psi$. The fuzzy function acts like a non-linear link function by taking in the input vectors $\boldsymbol{X}_N$ and transforming it into a value in the range [0,1] using the pre-specified rule base and the corresponding parameters $\theta$. The rule bases are specified as before (or can be learned using the rule base selection parameter) and the parameters of the fuzzy function can be estimated using the MCMC algorithm.   

As an illustrative example, we use the same rule base as in Eq. \ref{eq:rule123} where the input co-variates are in the range [0-10], but the output is in the range [0,1] using triangular membership functions as in Eq. \ref{eq:triang_mem}. We generate 100 data points from this fuzzy function and set any data point above a threshold of 0.5 to a value of 1 and any data point below it to a value of 0. This forms the training data-set for running the MCMC algorithm to see if the true parameters $\theta$ of the fuzzy membership function can be recovered which would classify the data-points into the two classes. Fig. \ref{fig:classifdata}, in the supplementary materials, shows the underlying surface generated with the true parameters $\theta$, with superimposed 100 randomly drawn samples which is used as the training data-set.

\begin{figure}[h!]
 \centering
 \includegraphics[width=0.99\columnwidth ]{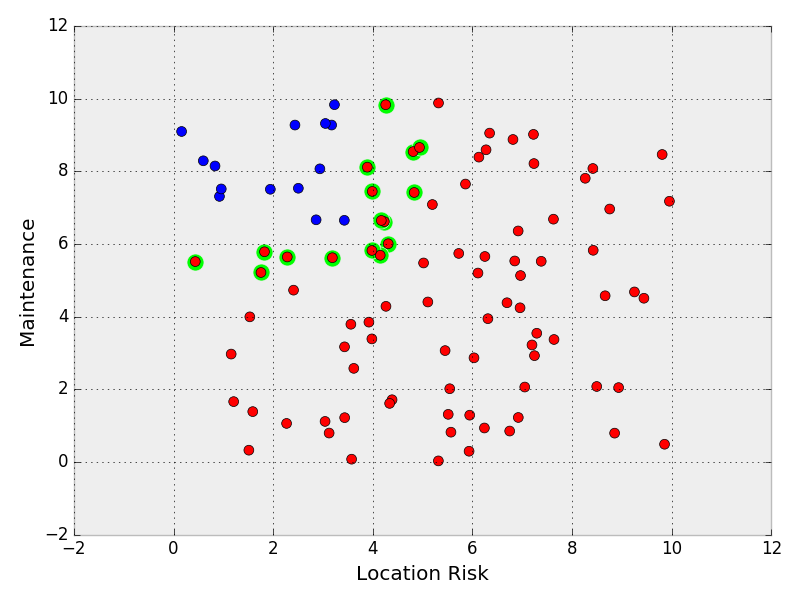}
 \caption{Binary classification results using the mean of the posterior samples in the Fuzzy Bayesian Learning approach. Predictions for Class 1 is in red, class 0 is in blue and the mis-classifications are shown with lime circles.}
 \label{fig:misclassif}
\end{figure}

For the simulations, 10,000 iterations are done for 3 independent chains and vague priors are chosen for $\theta$ as also done in all the regression cases. After rejecting 2000 samples for burn-in, the mean of the posterior parameter samples are plugged into the fuzzy function, with 0.5 as the threshold and the predicted class labels is plotted in Fig. \ref{fig:misclassif}.

One advantage of this method is that the introduction of the fuzzy system acts as a non-linear kernel and helps in classifying data which are not necessarily linearly separable in the feature space. Also, class one (in red) has much more number of data points than the other (blue) and therefore this is the so-called imbalanced data-set case \cite{he2009learning}. The method is able to determine the classes without any special modification for imbalanced data-sets (e.g. changing objective function values to penalise mis-classification of one class more than the other etc). However, as explained earlier in the regression cases as well, the caveat with this method is that the rule bases might not always be trivial to find and choosing a totally wrong rule base would drastically affect the performance of the classifier.

Also, the class label prediction plots in Fig. \ref{fig:misclassif} shows that there is a higher mis-classification at the class boundaries where the underlying fuzzy surface transits from greater than 0.5 to less than 0.5. This implies that the underlying fuzzy surface is not estimated properly with 100 data points, unlike the regression cases (with same or less number of data points) where the parameter estimates had really tight bounds. One possible reason might be that since the observations are just integer values (zeros or ones) instead of real numbers, there is less information for the MCMC algorithm to find the actual parameters with tighter uncertainty bounds.

\subsection{Application to a real world problem}
\subsubsection{Context and background of the problem}
The background of the problem is rooted in the financial services sector, specifically for making risk quantification models for the non-life speciality insurance domains. For this particular example, consider the case of insuring a power plant operation in the event of business interruption (BI) and property damage (PD) due to various risks like fire hazard, machinery breakdown etc. The focus of this paper is specifically the BI which depends on various factors like the reliability of each of the machinery components, redundancy schemes employed, maintenance schedules etc. Therefore, given a set of historic data on these co-variates and the corresponding system uptime in a year, it would be possible to fit a simple GLM or other regression models for prediction. The distribution of the system uptime can then be subsequently used in an insurance pricing model to obtain the required premium. This example focuses on better estimation of the system uptime for the insured engineering process.

It is possible to build more involved models of such engineering process systems in cases where individual failure and repair data of each of the underlying component is available as has been pursued in \cite{pan2016systems}. However, many plant operators are reluctant to share data and therefore such data might not be available for all different kind of process plants, to actually make a sizeable population so as to draw a statistical estimate of the failures and reliability of each component. The proposed Fuzzy Bayesian Learning (FBL) method is useful in such scenarios as it starts with some prior knowledge which informs the model parameters and some prior expert opinion which informs the model structure. As data comes in, the parameters can be re-estimated with these observations to get more tighter estimates around the model parameters and consequently predictions with less uncertainties. 

This case also serves as a validation of the ability to use this kind of modelling for other type of emerging risk areas where no prior insurance product has been offered and consequently there is little or no loss history data that can be used to build a data-driven model. In such cases, the fuzzy rule base relying on the underwriter's opinion without any data can be constructed to give a sense of the outputs. Later the Bayesian parameter estimation can be introduced as the actual claims data start coming in.

\subsubsection{Dataset description}
The data-set is collected in-house at Sciemus Ltd. London, for underwriting insurance risks for its power sector business. The data-set consists of 3 variables, which serve as an input to our model -- operation \& maintenance (O\&M), 
loss history, design \& technology \& fire risk (DnTnF). These are qualitative opinions indicated by the underwriter from the submission document of the specific risk which comes in to get a quotation for insurance.
All these variables are on a scale of 0-10 with 0 representing very poor and 10 indicating very good. Since the dataset consists of historical information from the actual underwriting process, it is possible to see which risk actually led to a loss event and a consequent claim, what was the claim amount and the amount of system uptime during the given insurance period. For this problem setting, the actual outcome is the the system uptime (percentage of time the system was working during the insurance period which is typically one year). This is a small data-set and Table \ref{table_summary_stats} gives the summary statistics of the 24 data-points available in it.

\subsubsection{Simulations}
\label{sec:simulations}

The fuzzy rule base used for simulation is shown in Table \ref{table_rule_base}. This is extracted from the intuition of the underwriter who believes that O\&M rating is the key to system uptime. This can be seen reflected in the table -- e.g. the first rule states that if the O\&M is good, then irrespective of the loss history and design factors the system uptime would be high. The rationale behind such a rule is that if the plant engineers know how the plant operates and are very rigorous with the maintenance schedule and monitoring aspects then they might get a few failures now and then (and have some history of losses), but this is inherently a good risk to underwrite. In some cases though, the loss history can be the overriding factor, e.g. rule 4 states that if the loss history is good, then the underwriter gives precedence to this over the bad O\&M. Similar lines of reasoning are extracted for the other rules as well. It is important to stress that multiple rules are triggered, each to a different extent, inside the fuzzy inference system, for one given antecedent input set. The final output is a combination of all these triggered values. 

\begin{table}[!th]
\renewcommand{\arraystretch}{1.3}
\caption{Rule base for the Fuzzy Bayesian model fitted on the real world case}
\label{table_rule_base}
\centering
\begin{tabular}{|c|c|c||c|}
\hline
\multicolumn{3}{|c||}{Antecedents}  & \multicolumn{1}{c|}{Consequent} \\
\hline
O\&M & Loss history &  DnTnF  & System uptime \\
\hline
\hline
Good & - & - & High \\
\hline
Good & - & Average & Medium \\
\hline
Average & - & - & High \\
\hline
Bad & Good & - & High \\
\hline
Bad & Bad & Bad & Low \\
\hline
\end{tabular}
\end{table}

Nevertheless, this rule base extraction is not exhaustive and neither all the possible model spaces are explored to see if they give similar results. A more rigorous approach to this would be Bayesian model comparison which involves comparing the evidence of different models to come up with the model that best explains the data. This is an investigation in its own right and a digression from the main theme of the paper which aims to introduce Fuzzy Bayesian Learning and therefore is not pursued in this paper. However, to show the effectiveness of this simple rule base, it is compared with the predictive accuracies of three GLMs (as shown in Eq. \ref{eq:GLM5}-\ref{eq:GLM7}) fitted with MCMC techniques. The generic likelihood as specified in Eq. \ref{eq:GLMgeneric} is used for all the GLMs.
The priors on each of the parameters are sufficiently non-informative to let the model learn from the dataset itself. The $\sigma$ in Eq. \ref{eq:GLMgeneric} is set to a fixed value of 0.25 (as also done while estimating the parameters of the Fuzzy Bayesian model). Ideally this should have been estimated from the data-set itself, but estimating the variance with tight uncertainty bounds actually requires large number of data points and hence this is pre-specified to a small value. Doing so implies that we are relying on the linear predictor to explain most of the relationships in the underlying data-set.  

For comparing among these models the mean squared error (MSE) is used to gauge the predictive accuracies. 
In Bayesian models, prediction of a new set of values $\tilde{\boldsymbol{y}}_{N}$  is done using the posterior predictive distribution:
\begin{equation}
\label{eq:posterior:predictive}
\begin{aligned}
p\left( \tilde{\boldsymbol{y}}_{N} | \boldsymbol{y}_{N} ,  \boldsymbol{X}_{N} \right)
&= \int{
p\left( \tilde{\boldsymbol{y}}_{N}, \theta | \boldsymbol{y}_{N} ,  \boldsymbol{X}_{N} \right)
d\theta}
\\
&= \int{
p\left( \tilde{\boldsymbol{y}}_{N}| \theta , \boldsymbol{y}_{N} ,  \boldsymbol{X}_{N} \right)
p\left( \theta | \boldsymbol{y}_{N} ,  \boldsymbol{X}_{N} \right)
d\theta}
\\
&= \int{
p\left( \tilde{\boldsymbol{y}}_{N}| \theta ,  \boldsymbol{X}_{N} \right)
p\left( \theta | \boldsymbol{y}_{N} ,  \boldsymbol{X}_{N} \right)
d\theta}
\text{.}
\end{aligned}
\end{equation}
Here,
$p\left( \tilde{\boldsymbol{y}}_{N}| \theta ,  \boldsymbol{X}_{N} \right)$
is the likelihood assumed on the model in Eq.~(\ref{eq:posterior}), and 
$p\left( \theta | \boldsymbol{y}_{N} ,  \boldsymbol{X}_{N} \right)$ is the posterior distribution.
Drawing samples from the posterior predictive distribution is simple when we already have draws from the posterior distribution as provided by the MCMC sampler \cite{gelman2014bayesian}.
We used the mean of the posterior predictive distribution, $\boldsymbol{y}_{N}^{\text{pred}} = \text{E}\left[\tilde{\boldsymbol{y}}_{N}\right]$ as our final prediction and compare it with the real data-point by calculating the mean squared error (MSE):
\begin{equation}
 \text{MSE} =
 \frac{1}{N}
 \left(\boldsymbol{y}_{N}^{\text{pred}}-\boldsymbol{y}_{N}\right)^{\text{T}}
 \left(\boldsymbol{y}_{N}^{\text{pred}}-\boldsymbol{y}_{N}\right)
\end{equation}
However, the MSE metric being a single number does not convey the uncertainties in the predictions properly, which can be gleaned from the whole posterior distribution of the prediction (but this again is difficult to compare with each other). The data-set is intentionally not divided into a training and a test set to perform hold out cross validation or k-fold cross validation, since this is a small data-set and ideally we would like the model not to lose out on much information so as to estimate the parameters with more certainty.

The resulting comparative analysis (using MSE as a metric), between the different models on the real world problem is shown in Table \ref{table_comparsion}. As can be observed, the Fuzzy-Bayesian model has a better MSE than GLM 5 which has a simple structure, but is worse off as compared to the other two GLMs (6 and 7). However, as discussed before, there is a narrative and more open interpretation for why the Fuzzy-Bayesian model works as opposed to the GLMs. Also predictive accuracy from such a small data-set is not really indicative of the performance of the model on the true underlying process and obtaining a lower MSE on this data-set might not always be the best option in the long run. As already discussed before, with new data points,  different GLMs might give lower MSEs. However, if the underlying process can be quantified with the fuzzy rule bases, then this would lead to a good fit in subsequent scenarios.

\begin{table}[!th]
\renewcommand{\arraystretch}{1.3}
\caption{Comparison of mean squared prediction errors for the fuzzy Bayesian model vis-a-vis other GLMs}
\label{table_comparsion}
\centering
\begin{tabular}{|c|c|}
\hline
Model & MSE \\
\hline
GLM 5 & 0.4098 \\
\hline
GLM 6 & 0.3649 \\
\hline
GLM 7 & 0.3596 \\
\hline
Fuzzy Bayesian & 0.3654 \\
\hline

\end{tabular}
\end{table}

The advantage of FBL in the real world example is that the underwriting insights can make the model better. But the other downside is that if the rules are extracted only from one underwriter, then their own biases and risk perceptions might adversely affect the model. To alleviate such concerns, the rule base can be constructed by consulting multiple underwriters and as more data comes in, the individual rule selection methodology (as shown in Section \ref{subsubsec:rule}) can be applied in the parameter estimation process to reject the rules that do not describe the data properly. This allows the model to leverage on the expertise of the best underwriters.

\section{Discussions and challenges}

\subsection{Relationship with other machine learning methods}

Most machine learning methods cannot intuitively explain the fitted model in terms of the underlying assumptions. For example, a neural network might give very good prediction accuracy on a data-set, but not much reasoning can be gleaned from the model, just by looking at the learned weights of the individual neurons. This might work in applications in an automated framework (like recommender systems, user behaviour prediction etc.) where large number of decisions need to be made in very short time scales without human interventions, and the financial liabilities associated with a mis-classification is not that high. However, in other industries like personalised healthcare/medicine or specifically our real world example of speciality non-life insurance, prediction from a computer model is almost always validated by a human before taking a final decision, since the consequences of such decisions have huge financial/legal implications. In such circumstances, having a deeper insight into why the algorithm actually works (as opposed to a black-box which just spits out gospel) makes it easier to take a decision and is always preferable. This issue is addressed by our proposed framework since the fuzzy rule base derived from the experts is used to inform the model structure for the regression or classification task. Also the associated uncertainties in the predictions would help the human decision maker to either ask for more information (i.e. recommend to do more medical tests for the healthcare case) or arrive at a different decision altogether due to the higher associated uncertainties (e.g. decide not to insure a specific risk or charge higher premiums to reflect the higher undertaking of risk).  

The other issue with traditional machine learning methods, is that they require quite significant amount of data to train and cross-validate a model properly. For the real world cases described above, even though the expert knows the dependence between the co-variates, it is almost always difficult and expensive to collect a comprehensive data-set and therefore modelling methodologies have to work with small data-sets and still give reasonable predictions. These kind of application areas would also be the ones where the proposed methodology would work well.  The advantage would also be that we can start with just the opinions of the expert without any data (there would be more uncertainties in the predictions) and then update the model with the data as we get more information (uncertainties in the predictions would decrease). The use of Fuzzy Bayesian Learning allows for such flexibility.

\subsection{Effect of informative priors}
In all the simulations, a have used noninformative priors. It might be especially useful to incorporate informative priors when there is even less data-points (e.g. underwriting emerging risks in a new sector which has only been done a few times or not at all). Judicious choice of the informative priors would help in giving smaller variance around the predictions. The predictive outputs for a Bayesian case is always a trade-off between the priors and the observed data. Therefore, as more data comes in, the model outputs are dictated by the data and the effect of these informative priors would diminish.

\subsection{Improvement over traditional GLMs}
This work can be seen as an extension of traditional GLMs with Bayesian parameter estimation. The proposed methodology retains all the advantage of Bayesian GLMs (e.g. simple model, intuitive explanation of parameters, inclusion of prior knowledge in the form of probability distribution), but makes the modelling methodology more flexible in the sense that the arbitrary choice of link functions can now be replaced with a model structure from the judgement of experts. Also qualitative imprecise opinions from the experts can be used to drive the model. Therefore, the model can handle both imprecision and uncertainty in the same framework.  

In fact, the general model structure is the same for our proposed model and a GLM.
Let $y$ be a response variable, and let $\boldsymbol{x}$ be a set of covariates observed alongside it.
Let $\mu$ and $\sigma^2$ be the mean and variance of a chosen error distribution, respectively.
We construct the general model structure as:
\begin{equation}
 \begin{aligned}
  y   &\sim \text{Error-Distribution}(\mu, \sigma^2)\\
  \mu &=    g(\boldsymbol{x}; \boldsymbol{\theta})
 \end{aligned}
\end{equation}
where $\boldsymbol{\theta}$ is the set of parameters to be estimated in the model.
For a GLM with $k$ covariates, we have:
\begin{equation}
  g(\boldsymbol{x}; \boldsymbol{\theta}) = h\left(\theta_0 + \sum_{i=1}^{k}{\theta_i x_i}\right)
\end{equation}
where $h(\cdot)$ is known as the inverse-link function, and it connects a linear combination of the parameters to the mean of the response variable.
In this paper, $g(\boldsymbol{x}; \boldsymbol{\theta})$ is the fuzzy system defined by the rulebase in Eq.~(\ref{eq:fuzzyrulebase}), so we replaced the inverse-link function and the linear predictor in the GLM.

\subsection{Relationship with type-2 and higher fuzzy sets}

A type-2 fuzzy set can incorporate uncertainty about its membership functions using a third dimension and its associated footprint of uncertainty (FOU). For interval type-2 fuzzy sets, the third dimension has the same value everywhere and can be completely defined by an upper \& lower membership function (UMF and LMF respectively) both of which are fuzzy type-1 sets. For this case, the area between the UMF and the LMF essentially represents the uncertainty. The Bayesian method can be used to learn from the data and reduce the area representing the uncertainties, thereby improving the predictions.

This framework therefore complements higher order fuzzy sets in the sense that irrespective of how the uncertainty is defined, the Bayesian method would act as a wrapper on top of the fuzzy logic based model, to reduce the band of uncertainty by learning from the data as more data-points come in. In terms of the present framework, defining an uncertainty range for the membership functions is somewhat similar to setting informative priors, instead of the uniform priors that we have considered in our simulations.  

The other key difference is that the fuzzy type-2 method  uses a type reducer between the inference and the de-fuzzification steps and the final output is always crisp, even though the membership functions have uncertainties in them. However, the proposed framework gives a posterior probability distribution for the predictions which helps in understanding the uncertainties associated with the predictions and aids in better decision making. It can also learn from new data that comes in and decrease the uncertainties in the membership functions and consequently in the predictions.

\subsection{Potential problems and unresolved issues with the methodology}
\label{potential_problems}
Since the model structure is shaped by the rule base provided by the experts, the predictive accuracy would be dependent on how good these rules are at explaining the data. Therefore, this methodology would work well in cases with small to medium data sets where the expert has good understanding of the problem domain. For other cases where there is a large number of data and there is not much domain knowledge, traditional machine learning methods like neural networks, support vector machines etc. would be a better alternative.

Proper choice of the specific rules would also affect the uncertainties in the parameter estimates and consequently in the predictions. Each rule would add an extra rule selection parameter ($\beta_k$) to be estimated by the MCMC from the data-set. If the data-set is small, estimation of more $\beta_k$'s indicate that the uncertainty in the parameter estimates would be higher. On the other hand if the rules are not exhaustive enough to explain the data, then the predictive accuracy of the model would be poor as the search space of the MCMC is effectively limited by this rule base set. Therefore, succinct and judicious choice of rules are important for smaller predictive uncertainties and better predictive accuracies.  

Even though the present work is geared primarily towards small and medium data-sets, the same methodology might be useful for large data-sets which do not fit in memory (i.e. big data). The rationale for this might be along similar lines to that of the present paper, where the interpretability of the model is also important vis-\'a-vis the predictive accuracy. However, Bayesian methods \& particularly MCMC is inherently a sequential algorithm and hence would not directly scale to huge datasets. However, partitioning the data and training the model on a subset of the data and recursively estimating the parameters by doing block updates might be a solution, i.e. the posterior distribution of the parameters estimated from one subset of the data becomes the prior distribution for the next subset. Other sampling methods or variational inference might also help in speeding up the computations, but more research needs to be done in this area. 

Including the fuzzy inference system and calling it at each MCMC iteration inherently makes the computation times much higher. Other solutions like developing surrogates for the expensive function and sampling from it only when required might be an avenue to explore. However, this might introduce additional errors in the parameter estimates as the surrogate model itself would be an approximation of the true function. Further research needs to be done on this.

\subsection{Directions for future research}

Future improvisations can look at recursive updating and sequential monte carlo methods (like particle filtering, unscented Kalman filtering etc.) so that when a new data-point arrives, the algorithm does not need to run again from scratch with the whole data-set and just updates the parameter values from the posterior of the old data-set and the new observation value. 

Another direction of research can be to look at more efficient sampling techniques than the present Metropolis-Hastings one, to improve convergence characteristics. Gradient free samplers like Slice sampling can be used for this case, but most other improved sampling like Hamiltonian Monte Carlo etc. would require gradients from the black box fuzzy function, which is difficult to obtain.  

Improvisations could also be made in the direction of computational intelligence, like extending the methodology to other kinds of fuzzy sets like intuitionistic fuzzy sets, non-stationary fuzzy sets, type-2 fuzzy sets, hesitant fuzzy sets etc. The fuzzy rule bases could be evolved through genetic programming or some variant of genetic algorithm \cite{cordon2001generating}, however this would potentially require more data for the learning process or might result in overfitting and consequent loss of generalisation capabilities of the model. Automated rule generation processes might also result in logically incoherent rule bases (even though they might have high predictive accuracy) and also loss of coherent reasoning behind the choice of such rules \cite{casillas2013interpretability}.

\section{Conclusion}
This paper introduced the symbiotic confluence of Fuzzy logic and Bayesian inference to learn from data and also incorporate the model structure in the form of an expert rule base. This helps in providing a narrative and transparency regarding the workings of the model while being able to include complicated non-linear transformations (in the form of rule bases) which give good predictive accuracy on a data-set. Unlike the generic fuzzy logic models, the outputs from the Fuzzy-Bayesian models are probabilistic and as demonstrated in the paper, can be used for regression, classification and also rule base selection tasks. Apart from synthetic examples, the approach is also demonstrated on a real life application in the financial services sector and is especially suited to small and medium data-sets. Extensive discussions are included for the relationship of this methodology with other machine learning methods and potential pitfalls like improper choice of rule-bases which would hamper the predictive capabilities of the model. A working example of the code is available at https://github.com/SciemusGithub/FBL to ensure reproducibility of the work, help readers build on it and implement their own ideas. Future directions for research involve using higher order fuzzy sets, improved sampling algorithms, recursive estimation and surrogate model based techniques for computational run-time reduction.


%
\section*{Acknowledgment}

The authors would like to thank Sciemus Ltd. for sponsoring this work. They would also like to thank Angelo Nicotra for helping with the underwriting information gathering process and Stefan Geisse for his insights in the power underwriting business.

\ifCLASSOPTIONcaptionsoff
  \newpage
\fi



%

\bibliographystyle{IEEEtran}
\bibliography{mybiblio.bib}

\clearpage

\begin{center}
\Large
Supplementary materials
\end{center}

\begin{figure}[b!]
 \centering
 \includegraphics[width=0.99\columnwidth ]{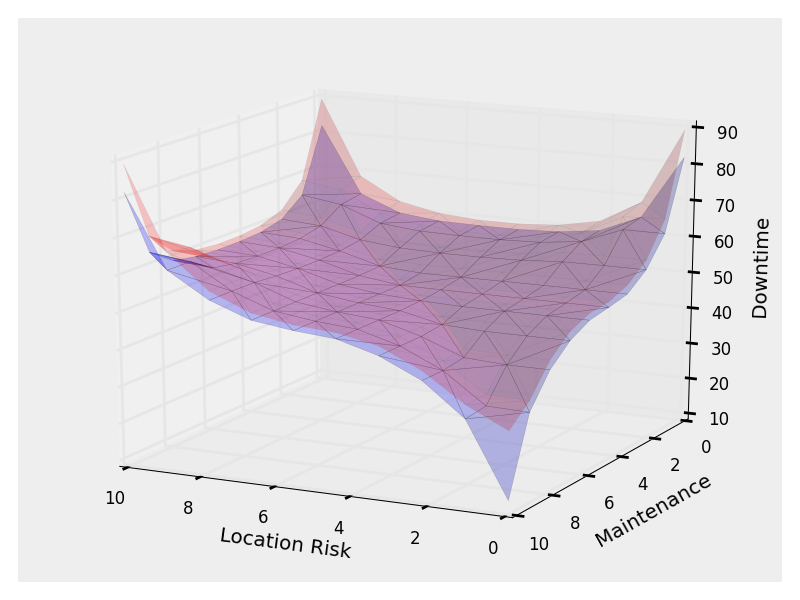}
 \caption{The envelope of the minimum and maximum values of Downtime for Location Risk and Maintenance obtained by running the fuzzy function with the parameter samples from the posterior}
 \label{fig:minmaxsurf}
\end{figure}

\begin{figure}[b!]
 \centering
 \includegraphics[width=0.99\columnwidth ]{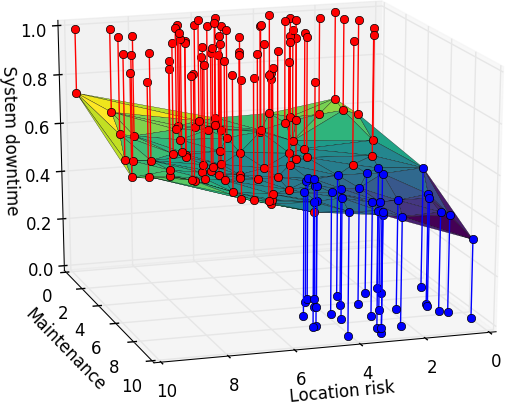}
 \caption{Underlying true surface from the fuzzy function which generates the two classes (red=1, blue=0) }
 \label{fig:classifdata}
\end{figure}

GLMs used for comparison with FBL in the real world case
\begin{equation}
\label{eq:GLM5}
GLM5: \mu= \sum_{i} \alpha_i x_{i}
\end{equation}
\begin{equation}
\label{eq:GLM6}
GLM6: \mu= \sum_{i} \alpha_i x_{i} + \sum_{\substack{
            i,j\\
            i \neq j}} \alpha_{ij} x_{i}x_{j}
\end{equation}
\begin{equation}
\label{eq:GLM7}
GLM7: \mu= \sum_{i} \alpha_i x_{i} + \sum_{\substack{
            i,j\\
            i \neq j}} \alpha_{ij} x_{i}x_{j}
            + \sum_{\substack{
            i,j,k\\
            i \neq j \neq k}} \alpha_{ijk} x_{i}x_{j}x_{k}
\end{equation}

\newpage

\begin{table*}[!th]
\renewcommand{\arraystretch}{1.3}
\caption{Summary statistics for input data}
\label{table_summary_stats}
\centering
\begin{tabular}{|c|c|c|c|c|}
\hline
& O\&M & Loss history &  DnTnF  & System uptime \\
\hline
count  & 24 & 24 & 24 & 24  \\
\hline
mean  & 6.71 & 6.42 & 5.71 & 8.54 \\
\hline
std  & 1.65 & 1.93 & 1.49 & 2.52 \\
\hline
min  & 4.00 & 2.00 & 2.00 & 2.00 \\
\hline
25\% & 6.00 & 5.00 & 5.00 & 7.75 \\
\hline
50\% & 7.00 & 6.50 & 6.00 & 10.00 \\
\hline
75\% & 8.00 & 7.00 & 6.25 & 10.00 \\
\hline
max  & 9.00 & 10.00 & 8.00 & 10.00 \\
\hline

\end{tabular}
\end{table*}

\begin{table*}[!htb]
\renewcommand{\arraystretch}{1.3}
\centering
\caption{Highest Posterior density estimates (L= Lower range of HDI, H=Higher range of HDI) for regression cases I-IV}
\label{regression_hdi}
\begin{tabular}{|c|c|c|c|c|c|c|c|c|c|c|c|}
\hline
                  &                   & \multicolumn{2}{c|}{Case I}              & \multicolumn{2}{c|}{Case II}             & \multicolumn{2}{c|}{\makeblue{Case III (a)}}  & \multicolumn{2}{c|}{\makeblue{Case III (b)}}            & \multicolumn{2}{c|}{Case IV}  \\
                  &                   & HDI (L)            & HDI (H)             & HDI (L)            & HDI (H)             & HDI (L)            & HDI (H)                  & \makeblue{HDI (L)}        & \makeblue{HDI (H)   }       & HDI (L)            & HDI (H) \\ \hline
                  &           Chain 1 & 4.997              & 5.002               & 5.000              & 5.000               & 4.761              & 5.620                    & \makeblue{4.708867}       & \makeblue{ 5.695282 }       & 4.587              & 5.476               \\
Param 1           &           Chain 2 & 4.997              & 5.003               & 5.000              & 5.000               & 4.768              & 5.600                    & \makeblue{4.710066}       & \makeblue{ 5.694645 }       & 4.592              & 5.468               \\
                  &           Chain 3 & 4.997              & 5.002               & 5.000              & 5.000               & 4.779              & 5.602                    & \makeblue{4.726268}       & \makeblue{ 5.664757 }       & 4.590              & 5.501               \\ \hline
                  &           Chain 1 & 4.998              & 5.003               & 4.999              & 5.000               & 4.650              & 5.676                    & \makeblue{4.528678}       & \makeblue{ 5.739515 }       & 4.528              & 5.482               \\
Param 2           &           Chain 2 & 4.998              & 5.003               & 5.000              & 5.000               & 4.657              & 5.686                    & \makeblue{4.533569}       & \makeblue{ 5.747654 }       & 4.469              & 5.502               \\
                  &           Chain 3 & 4.998              & 5.003               & 5.000              & 5.000               & 4.651              & 5.654                    & \makeblue{4.577323}       & \makeblue{ 5.797228 }       & 4.475              & 5.478               \\ \hline
                  &           Chain 1 & 4.999              & 5.001               & 5.000              & 5.000               & 4.948              & 5.661                    & \makeblue{4.892274}       & \makeblue{ 5.717771 }       & 4.608              & 5.303               \\
Param 3           &           Chain 2 & 4.999              & 5.001               & 5.000              & 5.000               & 4.906              & 5.614                    & \makeblue{4.855079}       & \makeblue{ 5.719516 }       & 4.641              & 5.369               \\
                  &           Chain 3 & 4.999              & 5.001               & 5.000              & 5.000               & 4.947              & 5.649                    & \makeblue{4.858045}       & \makeblue{ 5.679679 }       & 4.650              & 5.336               \\ \hline
                  &           Chain 1 & 4.999              & 5.001               & 5.000              & 5.000               & 4.188              & 4.964                    & \makeblue{4.167774}       & \makeblue{ 5.079541 }       & 4.603              & 5.480               \\
Param 4           &           Chain 2 & 4.999              & 5.001               & 5.000              & 5.000               & 4.198              & 5.002                    & \makeblue{4.114447}       & \makeblue{ 5.083357 }       & 4.553              & 5.415               \\
                  &           Chain 3 & 4.999              & 5.001               & 5.000              & 5.000               & 4.209              & 5.051                    & \makeblue{4.203955}       & \makeblue{ 5.102892 }       & 4.510              & 5.417               \\ \hline
                  &           Chain 1 & 4.997              & 5.002               & 4.999              & 5.000               & 4.682              & 5.523                    & \makeblue{4.612193}       & \makeblue{ 5.627037 }       & 4.573              & 5.412               \\ 
Param 5           &           Chain 2 & 4.998              & 5.002               & 5.000              & 5.000               & 4.650              & 5.543                    & \makeblue{4.620157}       & \makeblue{5.621127  }       & 4.539              & 5.389               \\
                  &           Chain 3 & 4.998              & 5.002               & 5.000              & 5.000               & 4.699              & 5.569                    & \makeblue{4.597553}       & \makeblue{ 5.543606 }       & 4.578              & 5.432               \\ \hline
                  &           Chain 1 & 4.999              & 5.001               & 4.999              & 5.001               & 4.205              & 5.222                    & \makeblue{4.128313}       & \makeblue{ 5.281103 }       & 4.509              & 5.485               \\
Param 6           &           Chain 2 & 4.999              & 5.001               & 4.999              & 5.001               & 4.205              & 5.241                    & \makeblue{4.116372}       & \makeblue{ 5.311240 }       & 4.481              & 5.442               \\
                  &           Chain 3 & 4.999              & 5.001               & 4.999              & 5.001               & 4.250              & 5.222                    & \makeblue{4.197869}       & \makeblue{ 5.326436 }       & 4.452              & 5.448               \\ \hline
                  &           Chain 1 & 49.922             & 50.060              & 49.970             & 50.031              & 41.959             & 72.715                   & \makeblue{40.127977}      & \makeblue{75.967792 }       & 35.040             & 83.099              \\
Param 7           &           Chain 2 & 49.927             & 50.063              & 49.975             & 50.034              & 39.990             & 71.748                   & \makeblue{40.221376}      & \makeblue{76.146952 }       & 33.714             & 82.078              \\
                  &           Chain 3 & 49.931             & 50.068              & 49.970             & 50.030              & 40.781             & 73.217                   & \makeblue{39.513554}      & \makeblue{75.267501 }       & 33.257             & 81.149              \\ \hline
                  &           Chain 1 & 49.995             & 50.006              & 49.998             & 50.002              & 50.788             & 54.042                   & \makeblue{50.544871}      & \makeblue{54.176391 }       & 48.119             & 51.368              \\
Param 8           &           Chain 2 & 49.995             & 50.006              & 49.999             & 50.002              & 50.785             & 53.874                   & \makeblue{50.486219}      & \makeblue{54.487353 }       & 48.499             & 51.657              \\
                  &           Chain 3 & 49.994             & 50.006              & 49.998             & 50.002              & 50.634             & 53.941                   & \makeblue{50.531675}      & \makeblue{54.252469 }       & 48.540             & 51.846              \\ \hline
                  &           Chain 1 & 49.964             & 50.055              & 49.992             & 50.008              & 53.843             & 65.629                   & \makeblue{52.555372}      & \makeblue{65.678843 }       & 41.053             & 57.158              \\
Param 9           &           Chain 2 & 49.963             & 50.051              & 49.992             & 50.008              & 53.852             & 65.324                   & \makeblue{52.118596}      & \makeblue{66.178505 }       & 41.318             & 57.918              \\
                  &           Chain 3 & 49.958             & 50.049              & 49.993             & 50.009              & 53.046             & 64.679                   & \makeblue{52.363778}      & \makeblue{66.301491 }       & 40.833             & 57.461              \\ \hline
                  & \makeblue{Chain 1}&                    &                     &                    &                     &                    &                          & \makeblue{0.980323 }      & \makeblue{ 1.316020 }       &                    &                     \\
\makeblue{Sigma}  & \makeblue{Chain 2}&                    &                     &                    &                     &                    &                          & \makeblue{0.977340 }      & \makeblue{ 1.312611 }       &                    &                     \\
                  & \makeblue{Chain 3}&                    &                     &                    &                     &                    &                          & \makeblue{0.970832 }      & \makeblue{ 1.302701 }       &                    &                     \\
\hline
\end{tabular}
\end{table*}

\begin{em} Cases for Tables \ref{regression_hdi} and \ref{regression_ess}: \end{em} \\
Case I: Regression with small dataset and no noise \\
Case II: Regression with bigger dataset and no noise \\
Case III: Regression with noise \\
Case IV: Regression with rule base selection \\
Note: Case IV also has the rule selection parameters which are binary and hence HDIs are not reported \\

\begin{table*}[!th]
\renewcommand{\arraystretch}{1.3}
\centering
\caption{Numerical convergence diagnostics for the regression cases I-IV, (ESS = Effective sample size, GL stat = Gelman Rubin statistic) }
\label{regression_ess}
\begin{tabular}{|c|c|c|c|c|c|c|c|c|c|c|c|}
\hline
                   & \multicolumn{2}{c|}{Case I}  & \multicolumn{2}{c|}{Case II} & \multicolumn{2}{c|}{\makeblue{Case III (a)}}  & \multicolumn{2}{c|}{\makeblue{Case III (b)}} & \multicolumn{2}{c|}{Case IV}  \\
                   & ESS   & GL stat              & ESS   & GL stat              & ESS   & GL stat                               &  \makeblue{ESS  }&  \makeblue{GL stat }      & ESS   & GL stat      \\ \hline
Param 1            & 26256 & 0.986                & 24922 & 0.995                & 22599 & 1.006                                 &  \makeblue{26846}&  \makeblue{1.011484}      & 25721 & 0.995        \\ \hline
Param 2            & 35320 & 0.964                & 24960 & 1.000                & 23192 & 0.997                                 &  \makeblue{29087}&  \makeblue{1.003060}      & 23239 & 0.999        \\ \hline
Param 3            & 21440 & 1.019                & 26320 & 0.990                & 24140 & 1.002                                 &  \makeblue{24893}&  \makeblue{1.025118}      & 26075 & 0.990        \\ \hline
Param 4            & 24326 & 0.997                & 26535 & 0.991                & 22565 & 1.007                                 &  \makeblue{26061}&  \makeblue{1.014798}      & 27549 & 0.990        \\ \hline
Param 5            & 40145 & 0.959                & 24867 & 0.998                & 24191 & 0.993                                 &  \makeblue{29637}&  \makeblue{0.999058}      & 25653 & 0.995        \\ \hline
Param 6            & 26620 & 0.992                & 25598 & 0.989                & 24643 & 0.996                                 &  \makeblue{28051}&  \makeblue{1.001591}      & 24795 & 0.998        \\ \hline
Param 7            & 41571 & 0.958                & 22331 & 1.012                & 23611 & 1.002                                 &  \makeblue{28616}&  \makeblue{1.002788}      & 25420 & 0.991        \\ \hline
Param 8            & 25070 & 0.995                & 27746 & 0.984                & 23939 & 1.004                                 &  \makeblue{23690}&  \makeblue{1.029275}      & 27761 & 0.992        \\ \hline
Param 9            & 43209 & 0.954                & 27110 & 0.985                & 25119 & 0.998                                 &  \makeblue{29310}&  \makeblue{1.001493}      & 26356 & 0.992        \\ \hline
\makeblue{Sigma}   &       &                      &       &                      &       &                                       &  \makeblue{18140}&  \makeblue{1.073940}      &       &              \\
\hline
\end{tabular}
\end{table*}

%








\end{document}